\documentclass[prd,11pt]{article}

\usepackage[utf8]{inputenc} 
\usepackage[T1]{fontenc}    
\usepackage{hyperref}       
\usepackage{url}            
\usepackage{booktabs}       
\usepackage{amsfonts}       
\usepackage{nicefrac}       
\usepackage{microtype}      
\usepackage{geometry}
\usepackage{authblk}
\usepackage{latexsym}

\usepackage{enumerate}
\usepackage[inline]{enumitem}
\usepackage{amsmath,amssymb}
\usepackage{amsfonts,dsfont}
\usepackage{nicefrac}
\usepackage{microtype}
\usepackage{mathtools}

\usepackage{sidecap}
\usepackage{caption}
\usepackage{subcaption}
\usepackage{wrapfig}
\usepackage{enumitem}
\usepackage[normalem]{ulem}
\usepackage{amssymb}
\usepackage{multicol}
\usepackage{adjustbox}
\usepackage{multirow}
\usepackage{color}
\usepackage{xspace}
\usepackage{CJKutf8}

\usepackage{mathpazo} 
\usepackage[scaled]{helvet} 
\usepackage{courier} 
\normalfont
\usepackage[T1]{fontenc}

\PassOptionsToPackage{numbers}{natbib}
\usepackage{natbib}
\usepackage{mkolar_definitions}
\usepackage{algorithm}
\usepackage[noend]{algpseudocode}

\newcommand\blfootnote[1]{%
  \begingroup
  \renewcommand\thefootnote{}\footnote{#1}%
  \addtocounter{footnote}{-1}%
  \endgroup
}

\newtheorem{assumption}{Assumption}

\newcommand{\alg}{SCC\xspace}
\newcommand{\hac}{HAC\xspace}
\newcommand{\dataset}{\ensuremath{\boldsymbol{X}}}
\newcommand{\tree}{\Tcal}

\newcommand{\defeq}{\vcentcolon=}

\newcommand{\clustering}{\mathbf{\mathbb{S}}}
\newcommand{\scc}{\textsc{SCC}\xspace}
\newcommand{\fc}{{e}\xspace}

\DeclarePairedDelimiter{\twoNormSqr}{\|}{\|_2^2}

\newcommand{\allclusterings}{\Pcal}

\newcount\Comments  
\definecolor{darkgreen}{rgb}{0,0.5,0}
\definecolor{darkred}{rgb}{0.7,0,0}
\definecolor{teal}{rgb}{0.1,0.6,0.7}
\definecolor{blue}{rgb}{0.0,0.1,0.9}
\definecolor{gblue}{rgb}{0.09, 0.35, 0.73}
\newcommand{\kibitz}[2]{\ifnum\Comments=1{{\textcolor{#1}{\textsf{\footnotesize [#2]}}}}\fi}

\AtBeginDocument{%
  \providecommand\BibTeX{{%
    \normalfont B\kern-0.5em{\scshape i\kern-0.25em b}\kern-0.8em\TeX}}}

\title{\bf Scalable Hierarchical Agglomerative Clustering}

\author{Nicholas Monath$^2$, Avinava Dubey$^1$, Guru  Guruganesh$^{1}$, Manzil Zaheer$^{1}$, \quad\quad Amr Ahmed$^{1}$,  Andrew McCallum$^{2}$, Gokhan Mergen$^{1}$, Marc Najork$^{1}$, \hspace{2cm} Mert Terzihan$^{4}$,  Bryon Tjanaka$^3$,  Yuan Wang$^{1}$,  Yuchen Wu$^{1}$ \\ $^1$ Google LLC \quad $^2$ University of Massachusetts Amherst \\ $^3$ University of Southern California  \quad
$^4$ Facebook } 
\date{\vspace{-1.7cm}}
\begin{document}
 
 
\maketitle

\blfootnote{Work done while NM and BT were interns at Google. Work done while MT was at Google. Corresponding author emails: \texttt{nmonath@cs.umass.edu} and \texttt{avinavadubey@google.com}.\\ Published DOI: \href{https://doi.org/10.1145/3447548.3467404}{\texttt{doi.org/10.1145/3447548.3467404}}}
\begin{abstract}
The applicability of agglomerative clustering, for inferring both hierarchical and flat clustering, is limited by its scalability. 
Existing scalable hierarchical clustering methods sacrifice quality for speed and often lead to over-merging of clusters.
In this paper, we present a scalable, agglomerative method for hierarchical clustering that does not sacrifice quality and scales to billions of data points.
We perform a detailed theoretical analysis, showing that under mild separability conditions our algorithm can not only recover the optimal flat partition, but also provide a two-approximation to non-parametric DP-Means objective \citep{jiang2012small}. 
This introduces a novel application of hierarchical clustering as an approximation algorithm for the non-parametric clustering objective. 
We additionally relate our algorithm to the classic  hierarchical agglomerative clustering method.
We perform extensive empirical experiments in both hierarchical and flat clustering
settings and show that our proposed approach achieves state-of-the-art results
on publicly available clustering benchmarks. Finally, we demonstrate
our method's scalability by applying it to a dataset of 30 billion queries. 
Human evaluation of the discovered clusters show that our method finds better quality of clusters than the current state-of-the-art. 
\end{abstract}
\section{Introduction}

Clustering is widely used for analyzing and visualizing large datasets
 (e.g, single cell genomics \cite{Schwartz2020-da} users in social networks \cite{Benevenuto2012-ml}), for solving tasks 
 such as entity resolution \cite{green2012entity,Zhang2018-pf,chen2018unique}, and for feature extraction (in debating systems \cite{ein-dor-etal-2018-learning} and knowledge base completion \cite{das2020probabilistic}). 
 While it is the case that many clustering tasks are NP-hard \cite{dasgupta2016cost,mahajan2009planar} and impossible to satisfy three simple properties \cite{kleinberg2002impossibility}, clustering is widely used and beneficial to the aforementioned applications in practice.

Hierarchical clustering, in which the leaves 
correspond to data points and the internal nodes correspond to clusters
of their descendant leaves, 
can be useful to represent 
clusters of multiple granularity \cite{zhang2014taxonomy} or to automatically
discover nested structures \cite{cranmer2019toy,gavryushkin2016space}.
Hierarchical clusterings represent multiple
alternative \emph{tree consistent partitions}  
\cite{heller2005bayesian}. Each tree consistent partition
is a set of internal nodes that correspond to a flat clustering 
of the dataset. This illustrates how hierarchical
clusterings can be used to represent uncertainty about
a candidate flat clustering. 
Hierarchical clustering methods are often used
to produce flat clustering, with a partition selected from
relevant nodes from the tree structure. 
This is commonly done in entity resolution \cite{green2012entity,Zhang2018-pf}. 
Extracting a flat clustering
from a hierarchical clustering, rather than
directly performing flat clustering, has been show to be empirically \cite{green2012entity, kobren2017hierarchical} as well as theoretically \cite{grosswendt2019analysis} beneficial.
The hierarchical structure has also has proved useful in interactive settings where user feedback is provided to improve and extract a flat clustering \cite{kobren2019integrating,vitale2019flattening}. 

Best-first, bottom-up, hierarchical agglomerative clustering (\hac) is
one of the most widely-used clustering algorithms \cite{eisen1998cluster,lee2012joint,green2012entity,Schwartz2020-da}.
It is used as the basis for inference in many statistical models \cite{blundell2010bayesian,heller2005bayesian,heller2005randomized}, as an approximation algorithm for hierarchical clustering costs \cite{dasgupta2016cost,moseley2017approximation} as well as for supervised clustering \cite{dean2018resolving,pmlr-v97-yadav19a}.
Interestingly, the hierarchical clustering algorithm has also been shown to be effective for flat clustering both theoretically in terms of K-means costs \cite{grosswendt2019analysis} as well as empirically \cite{green2012entity,kobren2017hierarchical}.
One capability that
contributes significantly to \hac's prevalence is that it can be used
to construct a clustering according to any cluster-level scoring
function, also known as a \emph{linkage function} \cite{dean2018resolving,pmlr-v97-yadav19a}. 

A key challenge in hierarchical clustering is scalability. 
For example, HAC takes $O(N^2\log(N))$ time for $N$ points.
Competing methods attempt to achieve better scalability by operating in an online manner \citep{monath2019scalable}. These incremental/online algorithms, while often effective
empirically,
are inherently sequential and so cannot utilize parallelism or scale to
datasets larger than a few million points \cite{monath2019scalable}. On the other hand,
randomized algorithms \citep{heller2005randomized} and parallel/distributed methods \citep{bateni2017affinity} typically achieve scalability at the cost of accuracy.

Affinity clustering \citep{bateni2017affinity}, overcomes the main computational expense in HAC algorithm 
by leveraging distributed connected component algorithms. 
While efficient and a state-of-the-art method, Affinity clustering suffers from over-merging clusters as we empirically observe in this work.  

In this paper, we design an accurate and scalable, bottom-up hierarchical clustering algorithm, the \emph{Sub-Cluster Component Algorithm (SCC)}. 
We provide a detailed theoretical and empirical analysis of in terms hierarchical clustering as well as flat clustering. 
We make the following contributions:\\

\noindent \textbf{Theoretical Contributions (\S \ref{sec:analysis})}
\begin{itemize}[topsep=0pt,itemsep=-1ex,partopsep=1ex,parsep=1ex]
    \item \alg produces hierarchies containing the optimal flat partition for data that satisfies {$\delta$-separability} \citep{KSB16} and achieves a constant factor approximation of \emph{DP-means objective} \citep{broderick2013mad}, a flexible objective for flat clustering which adapts to different numbers of clusters. This is to our knowledge the first use of hierarchical clustering to provide an approximation algorithm for this non-parametric objective
    \item\alg generalizes HAC (in limit we can recover the same tree as HAC). We also show that \scc can recover the target partition of model-based separated data \cite{monath2019scalable}, which is recoverable by HAC.
\end{itemize}

\noindent \textbf{Empirical Highlights (\S \ref{sec:exp_bench} \& \S\ref{exp:user_query})}
 \begin{itemize}[topsep=0pt,itemsep=-1ex,partopsep=1ex,parsep=1ex]
    \item State-of-the-art hierarchical and flat clustering results on publicly available benchmark datasets.
    \item Produces lower-cost clusterings in terms of DP-Means~\citep{pan2013optimistic} than competing state-of-the-art methods.
    \item Web scale experiments examining the coherence of the clusters discovered by our algorithm, on \textbf{30 billion user queries}. Human evaluation shows \scc produced more coherent clusters than Affinity clustering. To the best of our knowledge, this is the largest evaluation of clustering algorithms, there by showcasing the scalability of \scc.
\end{itemize}

\section{Sub-Cluster Component Algorithm}
\label{sec:scc}

In this section, we first formally define notation and then describe our proposed \scc algorithm.
 
\subsection{Definitions} 

Given a dataset of points $\dataset = \{x_i\}_1^N$, a flat clustering or partition is a set of disjoint subsets. Formally:
 
\begin{definition}\textbf{\emph{[Flat clustering]}}.
A {flat clustering} or partition of $\dataset$, denoted $\clustering = \{C_1,\dots,C_K\}$, of $\dataset$, is a set of disjoint (and non-empty) subsets (i.e., $C_i \cap C_j = \emptyset,\ \forall C_i \not = C_j$) that covers $\dataset$ (i.e., $\bigcup_{i=0}^K C_i = \dataset$).
\label{def:fc}
\end{definition}
 
\noindent  We refer to the set of all partitions of a set $\dataset$ as $\allclusterings(\dataset)$. Each flat clustering is a member of this set, $\clustering \in \allclusterings(\dataset)$. A hierarchical clustering is a recursive 
partitioning of dataset $\dataset=\{x_i\}_1^N$ into a tree-
structured set of nested partitions $\tree$. Formally: 

\begin{definition}\textbf{\emph{[Hierarchical clustering \citep{krishnamurthy2012efficient}]}}.
A hierarchical clustering, $\tree$, of a dataset $X={x_1,x_2,\dots,x_N}$, is a set of clusters $C_0 = \{x_i\}_{i=1}^N $ and for each $C_i, C_j \in \Tcal$ either $C_j \subset C_k$, $C_k \subset C_j$ or $C_j \cap C_k = \emptyset$. For any cluster $C \in \tree$, if $\exists C'$ with $C' \subset C$, then there exists a set $\{C_j\}_{j=1}^\ell$ of disjoint clusters such that $\bigcup_{i=1}^\ell C_j = C$. 
\label{def:hc}
\end{definition}

\noindent Equivalently, a hierarchical clustering can be thought of as a tree structure such that leaves correspond to individual data points and internal nodes represent the cluster of their descendant leaves. In this work, we will refer to nodes of the tree structure by the cluster of points that the node represents, $C_i$,  as in Definition \ref{def:hc}. Parent/child edges in the tree structure can be inferred using this cluster-based notation. A hierarchical clustering encodes many different flat clusterings, known as \emph{tree consistent partitions} \citep{heller2005bayesian}. A tree consistent partition is a set of internal nodes $\{C_1, \dots C_K\} \subset \Tcal$ that form a flat clustering.

Following HAC and previous work \cite{monath2019scalable,bateni2017affinity}, our algorithms will make use of \emph{linkage functions}, which measure the dissimilarity of two sets of points: $d: \mathcal{P}({\dataset}) \times \mathcal{P}({\dataset}) \rightarrow \RR^{+}$. Linkage functions are quite general in that they can 
support any function of the two sets \cite{monath2019scalable,lee2012joint,dean2018resolving}. Many commonly 
used linkage functions are defined in terms pairwise dissimilarities between points. For instance, the well-known
\emph{average} linkage is the average pairwise dissimilarities between points of one set to the other and \emph{single} is the minimum  pairwise dissimilarity. We will show how particular linkage functions 
can be used to achieve theoretical guarantees about our algorithm's performance.

\begin{figure*} 
    \centering
    \includegraphics[width=\textwidth]{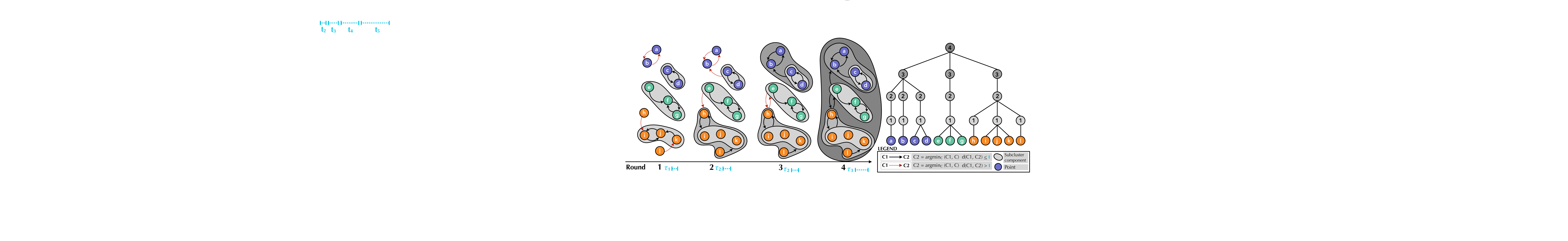} 
        \caption{\textbf{The Sub-Clustering Component Algorithm}. We illustrate SCC on a small dataset. The formation of sub-clusters
        is shown with black arrows for pairs of points satisfying Def.~\ref{def:sc}. The direction indicates
        indicates the nearest neighbor relationship (Def.~\ref{def:sc}, condition 2). Red edges indicate the nearest neighbor relationships that are above the distance thresholds. The grey circles indicate the sub-cluster components created in that round. Best viewed in color. 
    \label{fig:phca_algo_data}}
\end{figure*}

\subsection{Proposed Approach}
\scc works in a best-first manner: determining which points should 
belong together in clusters in a sequence of rounds. 
The sequence of rounds begins with the decisions that are ``easy to make'' (e.g., points that are clearly in the same cluster) and prolongs the later, more difficult decisions until these confident decisions have been well established. 
\scc starts by putting each point into its own separate cluster. 
Then in each round, we merge together groups of clusters from the previous round that satisfy a given certain ``condition''. The merging operation continues, until there are no pairs of clusters remaining to be merged.
Each round of our algorithm produces a partition (flat clustering) of the dataset at a different granularity and the collection of rounds together forms a hierarchical clustering (with non-parametric branching factor).

Let $\clustering^{(i)}$ be the partition produced after round $i$ and the partition at the starting round be $\clustering^{(0)} = \{\{x\}| x \in \dataset\}$. 
Let a \emph{sub-cluster} refer to a member a partition, i.e. $C \in \clustering^{(i)}$. 
Let $\tau_1,\dots \tau_L$ be a series of $L$ predefined increasing thresholds, given as hyperparameters to the algorithm.
To specify the ``condition" under which we merge sub-clusters in each round, we define \emph{sub-cluster component} as:

\begin{definition}{\textbf{\emph{[Sub-cluster Component]}}}
\label{def:sc}
Two sub-clusters 
$C_j, C_k \in \clustering$ are defined to be part of the same \emph{sub-cluster component} according to a threshold $\tau$ and linkage $d: \mathcal{P}({\dataset}) \times \mathcal{P}({\dataset}) \rightarrow \RR^{+}$, denoted $\textsc{Ch}_d(C_j, C_k, \tau, \clustering) = 1$, if   there exists a path $P \subseteq \clustering$ defined as $\{ C_j =  C_{s_0}, C_{s_1}, C_{s_2}, \ldots C_{s_{R-1}},  C_{s_R} = C_k  \}$, where each the following two conditions are met:
\begin{enumerate}
    \item  $d(C_{s_r}, C_{s_{r-1}}) \leq \tau \;\text{ for } \, 0 \leq r\leq R$, and
    \item either $C_{s_{r-1}} = \argmin_{C \in \clustering} d(C_{s_r}, C)$ and/or \\
     $C_{s_{r}} = \argmin_{C \in \clustering} d(C_{s_{r-1}}, C)$.
\end{enumerate}

\end{definition} 

Inference at round $i$ works by merging the sub-clusters in round $i-1$ that are in the same \emph{sub-cluster component}.
Computationally, the construction of sub-cluster components can be thought of as the connected components of a graph with nodes as the sub-clusters from the previous round and edges between pairs of nodes that are nearest neighbors and are less that $\tau$ from one-another. 

We define, $\textsc{SC}_d(C_j, \clustering, \tau)$, as the union of all sub-clusters in $\clustering$ that are within the sub-cluster component of $C_j$, i.e.,
\begin{equation}
    \textsc{SC}_d(C_j,\clustering, \tau) \defeq  \bigcup\limits_{\substack{C \in \clustering,\\ \textsc{Ch}_d(C_j, C, \tau, \clustering) = 1}}
     C  
     \label{eq:cls_sca} 
\end{equation}
Thus, $\textsc{SC}_d(C_j,\clustering^{(i-1)}, \tau^{(i)})$ is a new cluster, created by taking a union of all clusters from round $i-1$ that are in the sub-cluster component of $C_j$.
We create $\clustering^{(i)}$, flat partition at round $i$, as the set of all of these newly found clusters:
\begin{equation}
    \clustering^{(i)} \defeq  \{ \textsc{SC}_d(C, \clustering^{(i-1)},\tau^{(i)}) \, | \, C \in \clustering^{(i-1)} \} \label{eq:cs_sca}
\end{equation}

We refer to our algorithm as the {\textbf{\underline{S}ub-\underline{C}luster \underline{C}omponent algorithm (\scc)}}. 
Alg. \ref{alg:scc} gives pseudocode for \scc. We only increment the threshold if no clusters are merged in the previous round i.e. $\clustering^{(i-1)} = \clustering^{(i)}$.  
The sub-cluster component in a particular round can be found  
efficiently using a connected components algorithm \citep{boruuvka1926jistem}. Any of the rounds can be used as a predicted flat clustering.
A hierarchical clustering is given by $\bigcup \scc(\dataset, d, \{\tau_1,\dots,\tau_L\})$, the union of the sub-clusters produced by all rounds. Figure \ref{fig:phca_algo_data} provides an illustration of the SCC algorithm and the sub-cluster formation.

\begin{algorithm}[t]
\caption{Sub-Cluster Component Alg. (\scc)}
\begin{algorithmic}[1]
\State{\textbf{Input:} $\dataset$: dataset , $d$: set dissimilarity, $\{ \tau_1, \ldots, \tau_L \}$: a set of thresholds in increasing order} 
\State{\textbf{Output:} $(\clustering^{(0)}, \clustering^{(1)}, \ldots)$: One flat partition per round}
\State $\clustering^{(0)} \gets \{\{x\}\ |\ x \in \dataset \} $
\State $\textsf{idx} \gets 1$, $i \gets 1$ 
\While{$\textsf{idx} < L$}
\State  Set $\textsc{SC}_d(C_i, \clustering^{(i-1)}, \tau^{(i)})$, $\forall C_i \in \clustering^{(i-1)}$,  (Eq. \ref{eq:cls_sca})
\State Set $\clustering^{(i)}$  (Eq. \ref{eq:cs_sca})
\State $\textsf{idx} \gets \textsf{idx} + \II[\clustering^{(i)} =\clustering^{(i-1)}]$ 
\State $i \gets i + \II[\clustering^{(i)} \neq \clustering^{(i-1)}]$
\State $\tau^{(i)} \gets \tau_\textsf{idx}$
\EndWhile
\State \textbf{return} $(\clustering^{(0)}, \ldots, \clustering^{(i-1)})$
\end{algorithmic}
\label{alg:scc}\vspace{-1mm}
\end{algorithm}

\section{Analysis\label{sec:analysis}}
 
\scc is a simple, intuitive algorithm for clustering that
has a number of desirable theoretical properties.
We provide theoretical analysis of \scc used for both 
flat and hierarchical clustering. We analyze the separability conditions under which \scc recovers the target clustering and connect these results to the DP-Means objective as well as hierarchical clustering evaluation measures.  Lastly, we
show that in the limit of number of rounds our method will produce the same tree structure as agglomerative clustering.

\subsection{Recovering Target Clustering}

Separability assumptions in clustering provide a mechanism
to understand whether or not an algorithm effectively 
and efficiently recovers cluster structure
under ``reasonable'' conditions.  
If we can define a \emph{center} for a cluster of points, then we can define $\delta$-separability, which expresses a ratio between 
the center-to-center dissimilarities and the point which is farthest 
from its assigned center.

\begin{assumption} \label{asmp:delta-sep}\emph{\textbf{($\delta$-Separability~\citep{KSB16})}}
We say that the input data $\dataset$ satisfies $\delta-$separation, with respect to some 
target  clustering $\clustering^\star = \{C_1, C_2, \ldots, C_k\}$ if there exists centers $c_1^\star, \ldots 
c_k^\star$ such that for all $i \neq j$ $\norm{c^\star_i - c^\star_j } \geq  \delta \cdot R$ where $R:=\max_{l \in [k]} \max_{x 
\in C_l} \norm{x - c_l^\star} $ . 
\end{assumption}

Each round of \scc produces a flat clustering of a dataset $\dataset$.
 We will show that if $\dataset$ satisfies the above $\delta$-separability assumptions, one of the rounds of \scc will in fact be equal to the target clustering for the dataset, $\clustering^\star$, corresponding to the separated clusters.
Formally, we make the statement:

\begin{theorem}\label{thm:opt}
Suppose the dataset $\dataset$ satisfies the $\delta$-separability 
assumption with respect to the target clustering 
$\clustering^\star = \{C^\star_1,\dots,C^\star_k\}$  for $\delta \geq \gamma $.  $SCC(\dataset,d,\{\tau_0, \dots, \tau_L\})$ is set of partitions produced by \scc (Alg. \ref{alg:scc}) with $d(\cdot, \cdot)$ as average
linkage and
geometrically increasing thresholds i.e.~$\tau_i = 2^i \cdot \tau_0$. The target 
clustering is equal to one of the clustering produced by one round of SCC, $\clustering^\star \in SCC(\dataset,d,\{\tau_0, \dots, \tau_L\})$, where $\gamma = 6 $ for all metrics and $\gamma = 30$ for the $\ell_2^2$ distance and $\tau_0 \leq \min_{x,x' \in \dataset^2} \norm{x-x'}$.
\end{theorem}

The proof of Theorem \ref{thm:opt} is given in the supplemental material (\S \ref{proof:thm2}). Intuitively, we prove that, for the aforementioned bounds on within/across cluster distances, a geometric series will include a threshold that is larger than largest within cluster distance and smaller than the closest across cluster distance between any two sub-clusters. Having such a threshold $\tau^\star$, we will have a round $i$ with a flat clustering, $\clustering^{(i)}$, equal to the target clustering $\clustering^\star$,  $\clustering^{(i)} = \clustering^\star$.

We also consider a more general class of separable data, model-based separation \cite{monath2019scalable}, which specifies
when a particular linkage function ``separates'' a dataset. 
In model-based separation, we view the dataset $\dataset$
as the nodes in an undirected graph with latent (unobserved) edges. The edges
of this graph provide connected components, which correspond exactly to a target clustering $\clustering^\star$.

\begin{assumption}\textbf{\emph{(Model-based Separation \cite{monath2019scalable})}} Let $G=(\dataset, E)$ be a graph. Let the function $d:\Pcal(\dataset) \times \Pcal(\dataset) \rightarrow \RR$ be a linkage function that computes the similarity of two groups of vertices and let $g: \Pcal(\dataset) \times \Pcal(\dataset) \rightarrow \{0,1\}$ be a function that returns 1 if the union of its arguments is a connected subgraph of $G$. Dataset $\dataset$ is model-based separated with respect to $f$ if:
{\small\begin{align*} 
    \forall C_0,\ C_1,\ C_2 \subseteq \dataset,\ g(C_0, C_1) > g(C_0, C_2) \implies d(C_0, C_1) < d(C_0, C_2).
\end{align*}}
The target partition, $\clustering^\star$, which is model-based separated, corresponds to connected components in $G$.
\label{def:hsep}
\end{assumption}

We show that for datasets that are model separated our algorithm will contain the target clustering, i.e.,

\begin{proposition}
\label{prop:mbs}
  Given a dataset $\dataset$ and a symmetric injective linkage function $f$ s.t. $\dataset$ is model-based separated with respect to $f$, let $\clustering^\star$ be the target partition corresponding to the separated data.
  There exists a $\tau_1,\dots,\tau_L$ such that the hierarchical clustering $\bigcup \scc(\dataset, \{\tau_1,\dots,\tau_L\},d)$  discovered by \scc contains the target partition $\clustering^\star$.
\end{proposition}

Please refer to the supplemental material for the proof (\S \ref{proof:mbs}).

\subsection{Relation to Nonparametric Clustering}
\label{sec:dpmeans}

Next, we analyze the performance of our algorithm with respect to nonparametric, flat clustering cost functions. Nonparametric clustering, where the  number of clusters is not known \emph{a priori} and must be inferred from the data, is useful for many clustering applications \citep{andrews2014robust,jiang2012small,kulis2012revisiting}. 
DP-means \citep{jiang2012small,kulis2012revisiting,broderick2013mad} is an example of a widely used nonparametric cost function, that is obtained from the small variance asymptotics of Dirichlet Process mixture models. 

\begin{definition}\textbf{\emph{[DP-Means \citep{jiang2012small}]}} Given a dataset $\dataset$, a partition $\clustering = \{C_1,\dots,C_K\}$, such that cluster $C_l$ has center $c_l$
and hyperparameter $\lambda$, the \emph{DP-Means} objective is:
\begin{equation}
    DP(\dataset, \lambda, \clustering) = \sum_{C_l \in \clustering} \sum_{x \in C_l} \norm{x - c_l}^2 + \lambda |\clustering|.
\end{equation}
Given a dataset $\dataset$ and hyperparameter $\lambda$, clustering according to DP-Means seeks to find: $\argmin_{\clustering,\textbf{c}} DP(\dataset, \lambda, \clustering)$ where $\textbf{c}$ are the centers for each cluster in $\clustering$.
\end{definition}

We show that \scc yields a constant factor approximation to DP-Means solution under $\delta$-separation (Assumption \ref{asmp:delta-sep}).

\begin{theorem}
\label{thm:dpmeans}
Suppose the dataset $\dataset$ satisfies the $\delta$-separability 
assumption with respect to the target clustering 
$\clustering^\star = \{C^\star_1,\dots,C^\star_k\}$  for $\delta \geq \gamma $.  $SCC(\dataset,d,\{\tau_0, \dots, \tau_L\})$ is set of partitions produced by \scc with $d(\cdot, \cdot)$ as the average
distance between points and
geometrically increasing thresholds i.e.~$\tau_i = 2^i \cdot \tau_0$. $SCC(\dataset,d,\{\tau_0, \dots, \tau_L\})$ contains a $ 2$-approximation solution to the DP-Means objective. 
\end{theorem} 
See Appendix \S \ref{proof:thm2} for the proof. The two primary steps are: (1)
    Using theorem \ref{thm:opt} and show that \scc can find optimal solution to the \emph{facility location} problem and (2) show that this solution is  a constant factor approximation of the DP-means solution.

Our analysis of \scc as an approximation algorithm for DP-Means shows the first connection of hierarchical bottom up clustering to non-parametric flat clustering objectives such as DP-means. This theoretical connection is also empirically useful as \scc achieves SOTA for the DP-means objective (\S \ref{sec:expt_dp}).

\subsection{Hierarchical Clustering Analysis}
\label{sec:hac}
 
\scc is reminiscent of hierarchical agglomerative clustering (HAC), which, in each round, merges the two subtrees with minimum distance according to the linkage function.
In the following statement, we show that there exists a sequence
of thresholds $\tau_0,\dots,\tau_L$, for which \scc will produce
exactly the same tree structure as HAC. 
To formally make this statement, 
we need to make the additional assumption that the linkage function is 
both injective (so as there are no ties in the ordering of mergers)
and reducible.
 
\begin{proposition}
Let $d: \mathcal{P}({\dataset}) \times \mathcal{P}({\dataset}) \rightarrow \RR^+$ be a linkage function that is symmetric and injective, $ C_1,C_2,C_3,C_4 \subset \dataset,\ d(C_1,C_2) = d(C_3,C_4) \iff (C_1 = C_3 \wedge C_2 = C_4) \vee (C_1 = C_4 \wedge C_2 = C_3)$.  Let $\tree$ be the tree formed by HAC and
let $d$ also satisfy reducibility, $\forall C,C',C'' \in \tree,\ f(C,C') \leq \min \{d(C,C''),d(C',C'')\} \implies \min\{d(C,C''), d(C',C'')\} \leq d(C \cup C', C'')$
then there exists a sequence of threshold $t_1,\dots,t_r$ such that  
the tree formed by SCC, i.e.,  $\bigcup \scc(\dataset, f, \{\tau_1,\dots,\tau_r\})$, is the same as $\tree$.
\label{prop:hac}
\end{proposition}

\textsc{Proof}.
Each node in the HAC tree, $C \in \tree$, has an associated linkage function score, denoted (with abuse of notation) as $d(C)$. We define the threshold-based rounds for \alg such that $t_1,\dots,t_r$ to be the values, $\{d(C) + \epsilon | C \in \tree \}$
sorted in ascending order. Because $f$ is reducible and injective, there is a unique pair of nodes that will be merged in each round. This pair will correspond exactly to the pair that is merged by HAC in the corresponding round. It follows that the resulting tree structures will be identical. $\square$

The connection between \scc and agglomerative clustering is used to prove Proposition~\ref{prop:mbs}. We note that \scc produces non-binary
tree structures. And so \scc is not meant to directly approximate the binary tree produced by HAC,  nor is it
not well suited for objectives such as Dasgupta's
cost \cite{dasgupta2016cost}.

\subsection{Time Analysis}
\label{alg:sparse}

\paragraph{Worst Case} Building the sub-cluster components
in each round is achieved by first finding the 1-nearest neighbor
of each point that is less than the given round threshold.
Then forming weakly connected components in the graph
with edges of the aforementioned 1-nearest neighbor relationships.
For a given round, let $O(T)$ be the time required
to build a 1-nearest neighbor graph over the sub-clusters
of a round. Cover Trees \citep{beygelzimer2006cover} allow
this to be done in $O(c^{12} N \log N)$ for $N$ elements where $c$ is the expansion constant. Let $O(S)$ be the time required to run connected components.
 Since we are finding
the connected components of a graph with $N$ nodes and at most $N$ edges, a worst case running time of connected components is $O(N)$. We note that each of the one nearest neighbor graph and the connected components
algorithm are highly parallelizable in a map-reduce framework.
In the worst case, \scc requires $2*(N-1)$ rounds (merging one pair of elements per round, multiplicative factor due to determining when to advance $\textsf{idx}$ in Algorithm~\ref{alg:scc}). In general this is a $O(N(T+S))$ running time.
We find that using a fixed number of rounds and simply advancing the threshold in each round experimentally works well.

\paragraph{Sparse Graphs}
To speed up computation, we can pre-compute a $k$ nearest-neighbor graph 
over the dataset. Once such a graph is constructed computing the 1-nearest 
neighbor of each node can be an $O(k)$ operation for many linkage functions
(such as average, single, complete). We can use the sparse $k$ nearest-neighbor
graph to compute ``top-k'' approximation of the linkage functions. 
For instance, to compute the average linkage between two clusters 
of points according to the $k$-nearest-neighbor graph, 
we can define the distance 
between any two points that do not share an edge in the sparse $k$ nearest neighbor
graph to be a fixed constant (e.g., 4 in the case of $\ell_2^2$ or 0 in the case 
similarities).

\section{Experiments}
\label{sec:exp_bench}

We empirically validate the effectiveness of \scc. 
Paralleling our theoretical contributions, we 
analyze the method both as hierarchical clustering approach as well as
a flat clustering method and DP-Means approximation algorithm. 
Analysis on a variety of publicly available clustering benchmarks demonstrates that \scc:
\begin{itemize}[topsep=0pt,itemsep=-1ex,partopsep=1ex,parsep=1ex]
    \item Recovers more accurate hierarchical clustering than state-of-the-art methods (\S \ref{sec:expt_hc}).
    \item Produces high quality flat partitions of the data (\S \ref{sec:expt_fc}).
    \item Produces lower DP-means cost solutions (\S \ref{sec:expt_dp}).
\end{itemize}

Lastly, to demonstrate the scalability of \scc we evaluate on \textbf{30 billion} point web-scale dataset  (\S\ref{exp:user_query}).

\subsection{Hierarchical Clustering \label{sec:expt_hc}}

\textbf{Datasets:}
We evaluate the \alg and competing methods on the following publicly available clustering benchmark datasets as in \citep{kobren2017hierarchical} (Table~\ref{tbl:dp_table}): 
\textbf{CovType }- forest cover types;
\textbf{Speaker} - i-vector speaker recordings, ground truth clusters refer each unique speaker \citep{greenberg2014nist};
\textbf{ALOI} - 3D rendering of objects, ground truth clusters refer to each object type \citep{geusebroek2005amsterdam};
\textbf{ILSVRC (Sm.) (50K subset)} and \textbf{ILSVRC (Lg.) (1.2M Images)} images from the ImageNet ILSVRC 2012 dataset \citep{russakovsky2015imagenet} with vector representations of images from the last layer of the Inception neural network.

\textbf{Methods} We analyze the performance of \alg compared to state-of-the-art hierarchical clustering algorithms:
\textbf{Affinity} \citep{bateni2017affinity} - a distributed minimum spanning tree approach based on Bor\r{u}vka's algorithm \citep{boruuvka1926jistem}; 
\textbf{Perch} \citep{kobren2017hierarchical} - an online hierarchical clustering algorithm that creates trees one point at a time by adding points next to their
nearest neighbor and perform local tree re-arrangements in the form of rotations;
\textbf{Grinch} \citep{monath2019scalable} - similar to \textsc{Perch} this is an online tree building method, this work uses a grafting subroutine in addition to rotations. The graft subroutine allows for more global re-arrangements of the tree structure; 
\textbf{gHHC} \citep{monath2019gradient} - a gradient-based hierarchical clustering method that uses a continuous tree representation in the unit ball; \textbf{hierarchical K-Means (HKM)} - the top down divisive algorithm; \textsf{hierarchical agglomerative clustering (HAC)} - the classic bottom up agglomerative approach;
\textbf{Birch} \citep{zhang1996birch} - a classic top-down hierarchical clustering method;
\textbf{HDBSCAN}  \cite{campello2013density} - the hierarchical extension of the classic DBSCAN algorithm. For {Perch}, {Grinch}, {HKM}, {Birch}, {gHHC} we report results from previous work \cite{kobren2017hierarchical,monath2019gradient,monath2019scalable}.
Cosine similarity is 
meaningful for each of the datasets. For all methods that use a linkage function, 
we use average linkage which was shown to be effective \cite{monath2019scalable}. 

\begin{table}[t]
\centering
\begin{tabular}{@{}l@{}rrrrrr@{}}
    \toprule
    & \bf \small{CovType} & \bf  \small{ILSVRC} &  \bf \small{ALOI} & \bf  \small{Spkr.}  & \bf  \small{ILSVRC} \\
    &  & \bf  \small{(Sm.)} &   & &  \bf  \small{(Lg.)} \\
    \midrule
    $|\clustering^\star|$  & 7 & 1000  & 1000  & 4958  & 1000 \\
     $|\dataset|$ & 500K & 50K & 108K & 36.5K   & 1.3M \\
     dim. & 54 & 2048 & 128  & 6388  & 2048 \\
     \midrule
     BIRCH & 0.44 & 0.26 & 0.32 & 0.22 &  0.11 \\
     \textsc{Perch} & 0.448 & 0.531 &0.445  &  0.372 & 0.207  \\
     HAC & - &   0.641 & 0.524 & 0.518  & - \\
     \textsc{Grinch} & 0.430 & 0.557 & 0.504 & 0.48 & - \\
     HKM  & 0.44& 0.12  & 0.44 & 0.12 & 0.11 \\
     gHHC & 0.444 & 0.381 & 0.462 & -  & 0.367 \\ 
     \textsc{HDBSCAN} & \bf 0.473 & 0.414 &0.599 & 0.396 & - \\
     Affinity & 0.433 & 0.587 & 0.478 & 0.424  & 0.601\\
     \midrule
     \alg & 0.433 & \bf  0.644 & \bf 0.619 & \bf 0.528 & \bf0.614 \\
     \bottomrule
\end{tabular}
\caption{\textbf{Dendrogram Purity} results on benchmark datasets.
gHHC did not produce meaningful results on Speaker and \textsc{Grinch} and \textsc{HDBSCAN} did not scale to ILSVRC (Lg.).} 
\label{tbl:dp_table}
\end{table}

\begin{figure*}[ht]
    \centering%
    \includegraphics[width=0.24\textwidth]{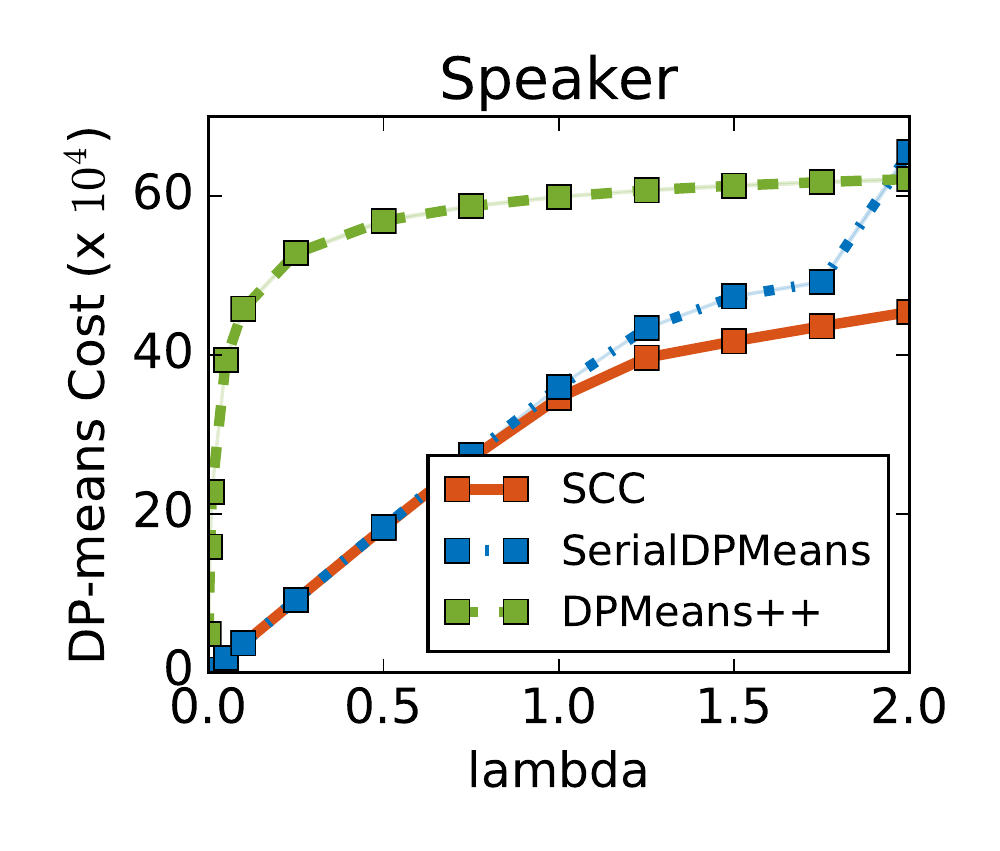}
    \includegraphics[width=0.24\textwidth]{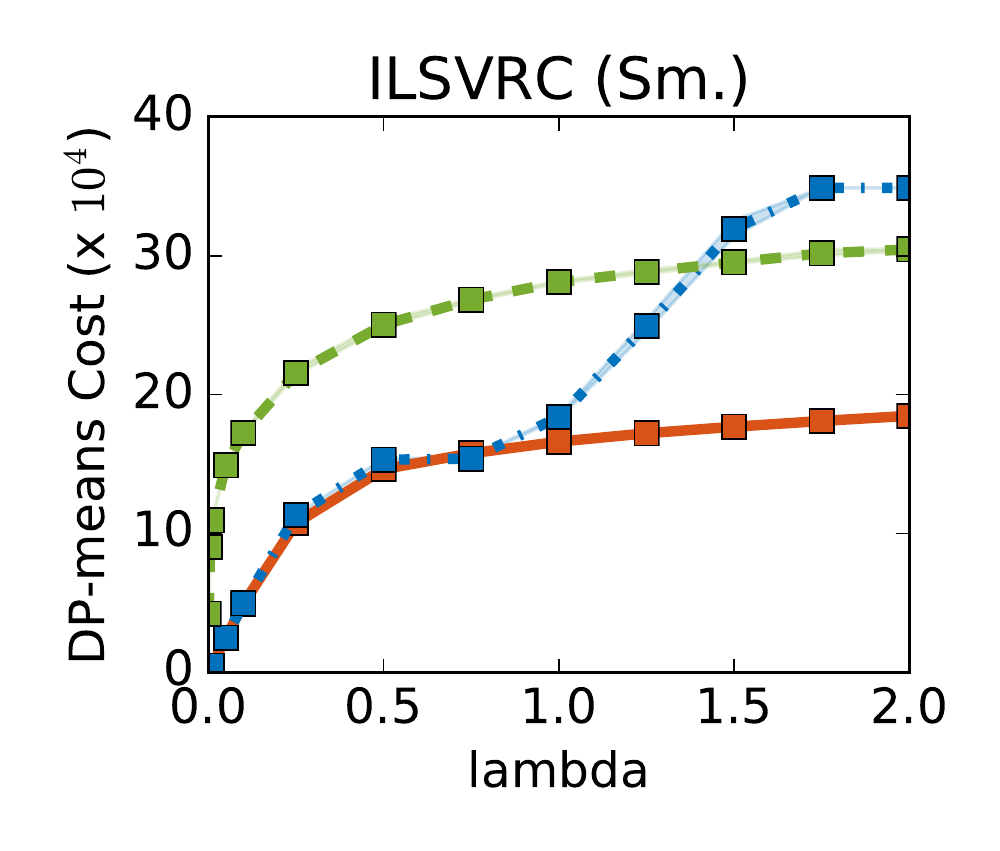}
    \includegraphics[width=0.24\textwidth]{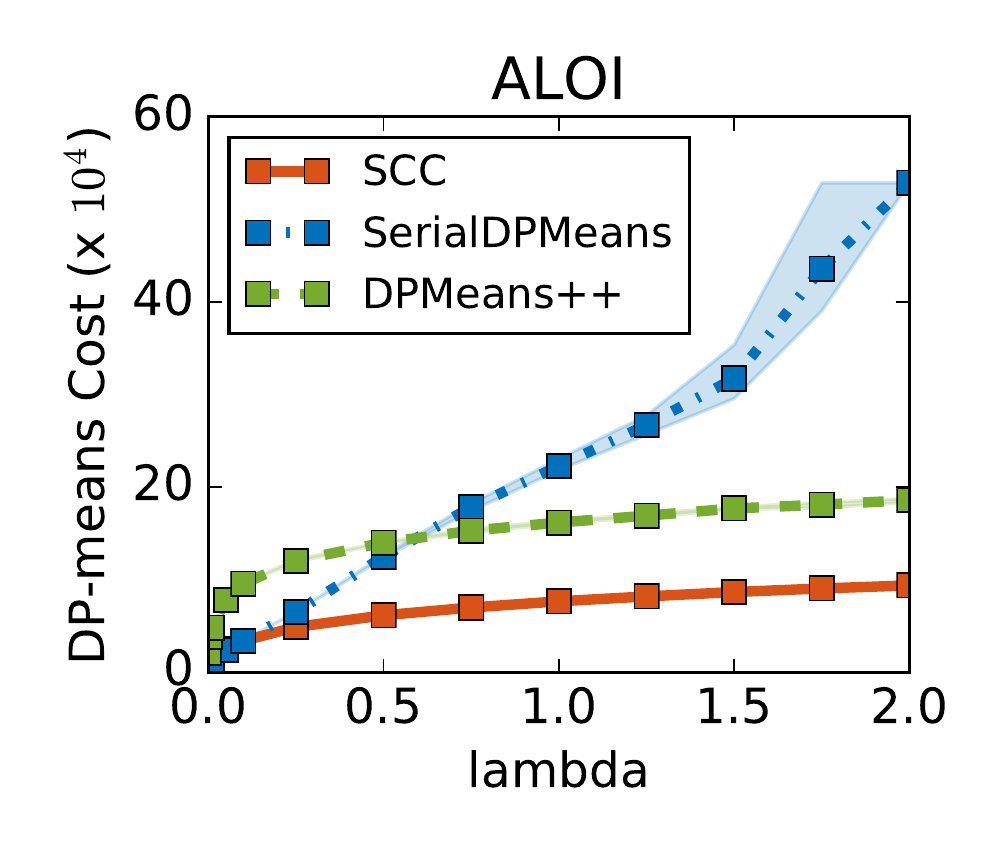}
    \includegraphics[width=0.24\textwidth]{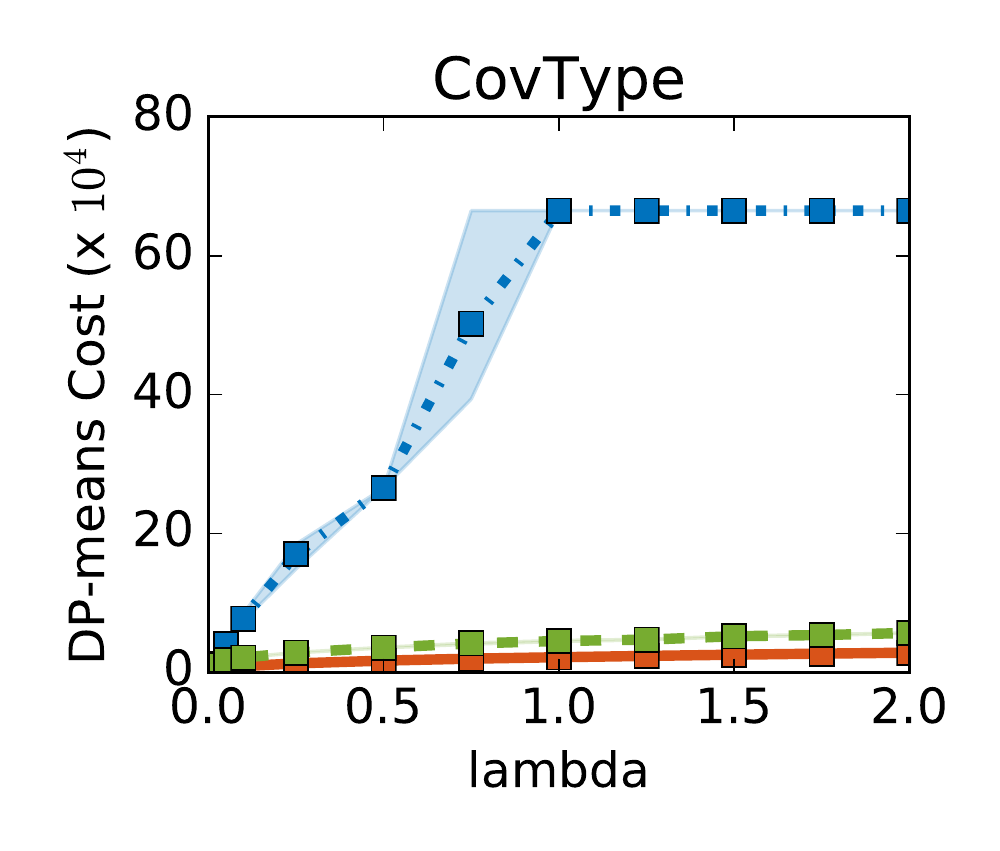}
    \vspace{-3mm}
    \caption{\textbf{DP-Means Cost} for the solutions found with a variety of methods for a range of $\lambda$ values from close to 0 to 2. \scc produces lower cost solutions for different values of $\lambda$ as compared to the other methods.}
    \label{fig:dpmeansplots}
\end{figure*}

\begin{figure*} 
    \centering%
    \includegraphics[width=0.24\textwidth]{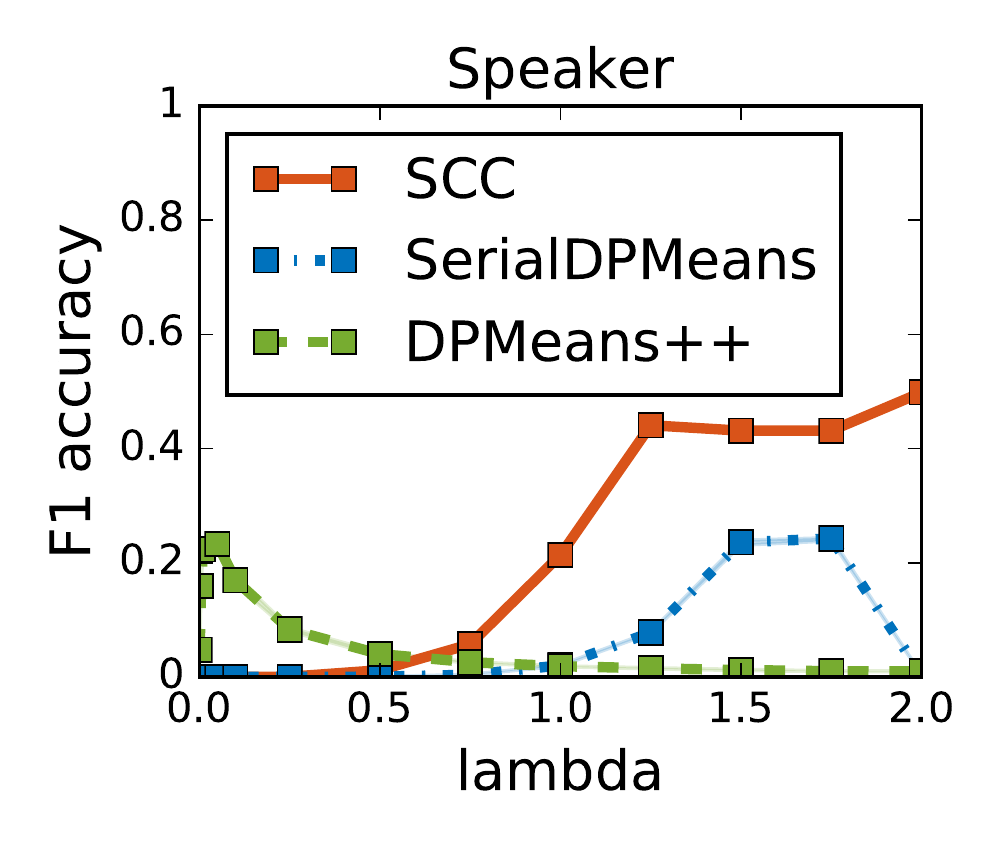}
    \includegraphics[width=0.24\textwidth]{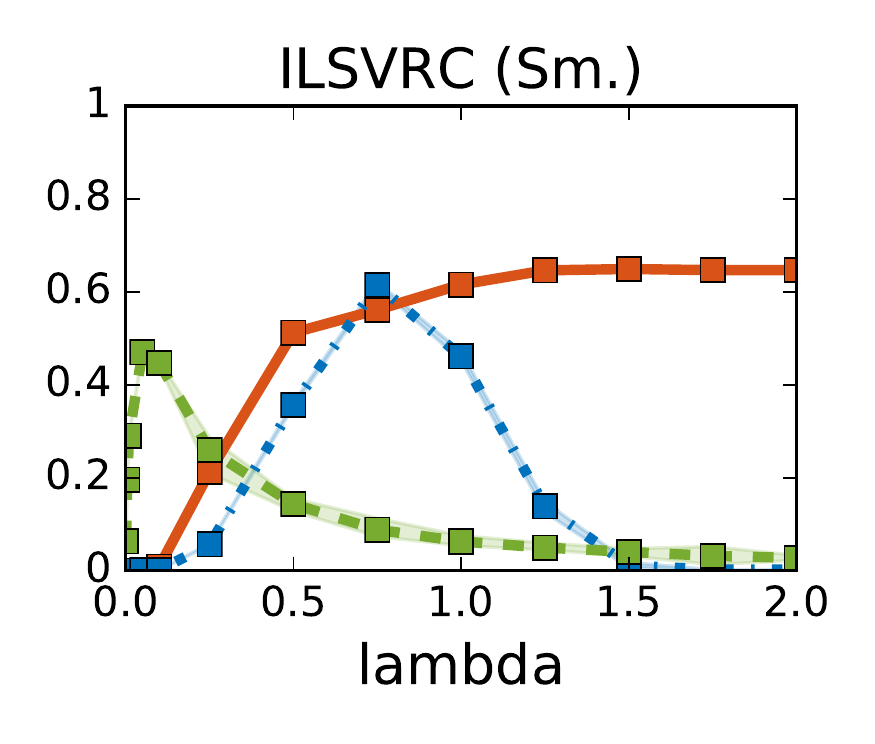}
    \includegraphics[width=0.24\textwidth]{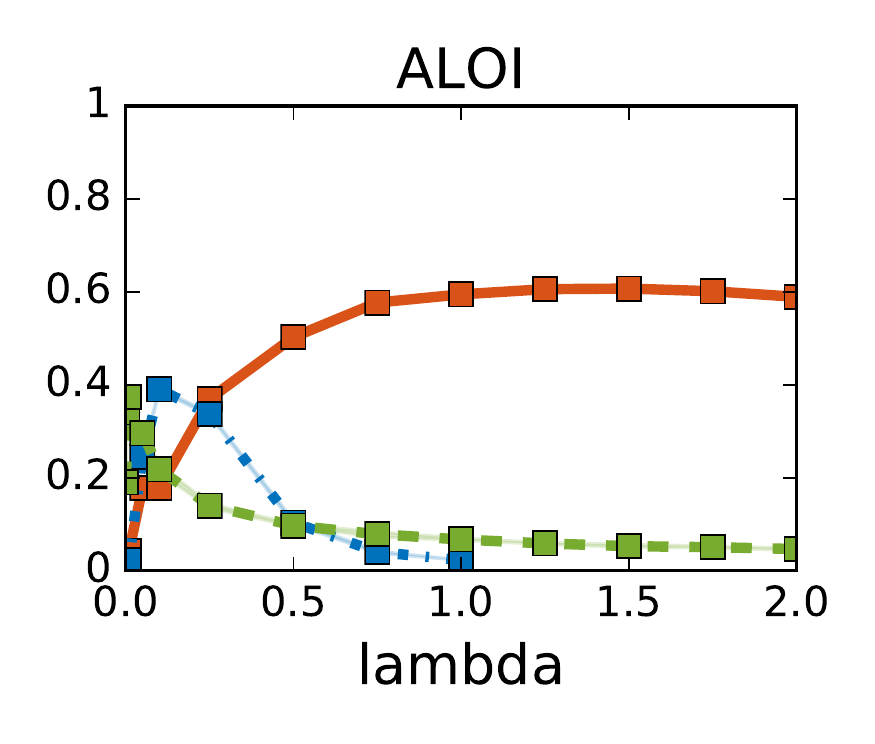}
    \includegraphics[width=0.24\textwidth]{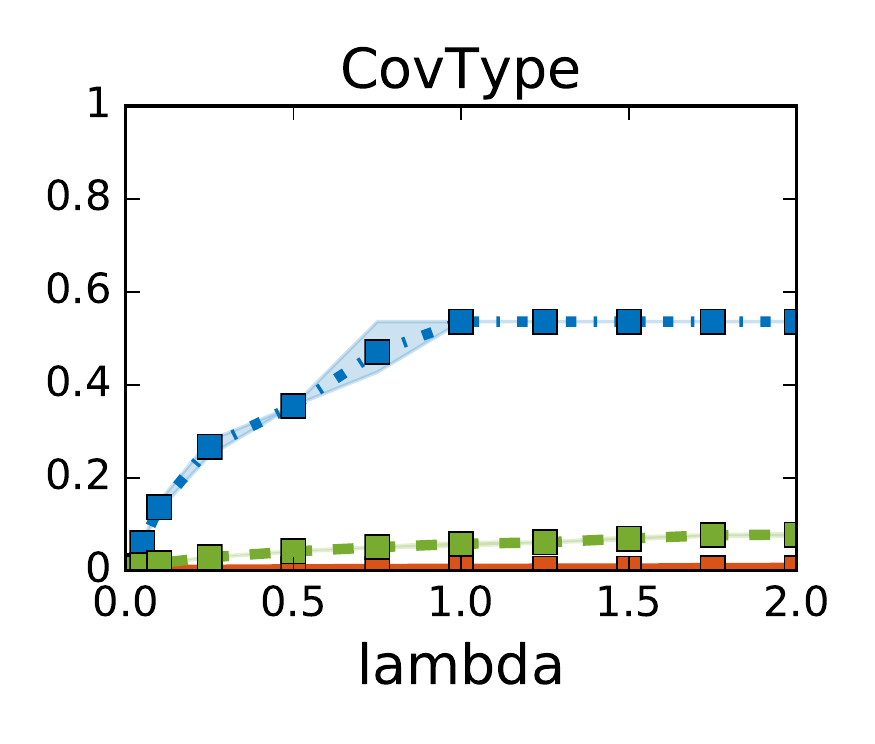}
        \vspace{-3mm}

        \caption{\textbf{Pairwise F1 Evaluation}.
        As each algorithm depends on $\lambda$ in a different way, the settings of $\lambda$ resulting in the best performance might differ between methods. We plot the performance of each method for each value of $\lambda$. When considering the best F1 achieved by each method for some value of $\lambda$, \scc is the top performing method on 4 of 5 datasets.
        }
    \label{fig:pairwise_eval} 
\end{figure*}

Since the datasets use cosine similarity, for SCC, we use a geometric progression of thresholds for \scc based on the bounds of cosine similarity (switching the sign of threshold condition), which we approximate with minimum (0.001) to maximum (1.0) with 200 thresholds for the algorithm. We use the sparsified nearest neighbor graph approach with $k$=25 neighbors using we  ScaNN \cite{avq_2020}.

Since each dataset has ground truth flat clusters, we evaluate the quality of hierarchy using \emph{dendrogram purity} as in previous work \cite{heller2005bayesian,kobren2017hierarchical,monath2019scalable}. 
 Dendrogram purity is the average, over all pairs of points from the same ground truth cluster, of the purity of the least common ancestor of the pair.
We hypothesize that SCC's improvement over Affinity clustering comes from its use of round-based
thresholds to reduce the overmerging of clusters.

As seen in Table~\ref{tbl:f1_all}, we observe that \alg achieves the highest dendrogram purity on all datasets except CovType. Notably, both \alg and Affinity clustering scale much 
better to the largest 1.2M image dataset, with both methods having no degradation in performance  from the 50K subset to the larger 1.2M point dataset. 

\begin{figure}
    \centering%
    \includegraphics[width=0.4\textwidth]{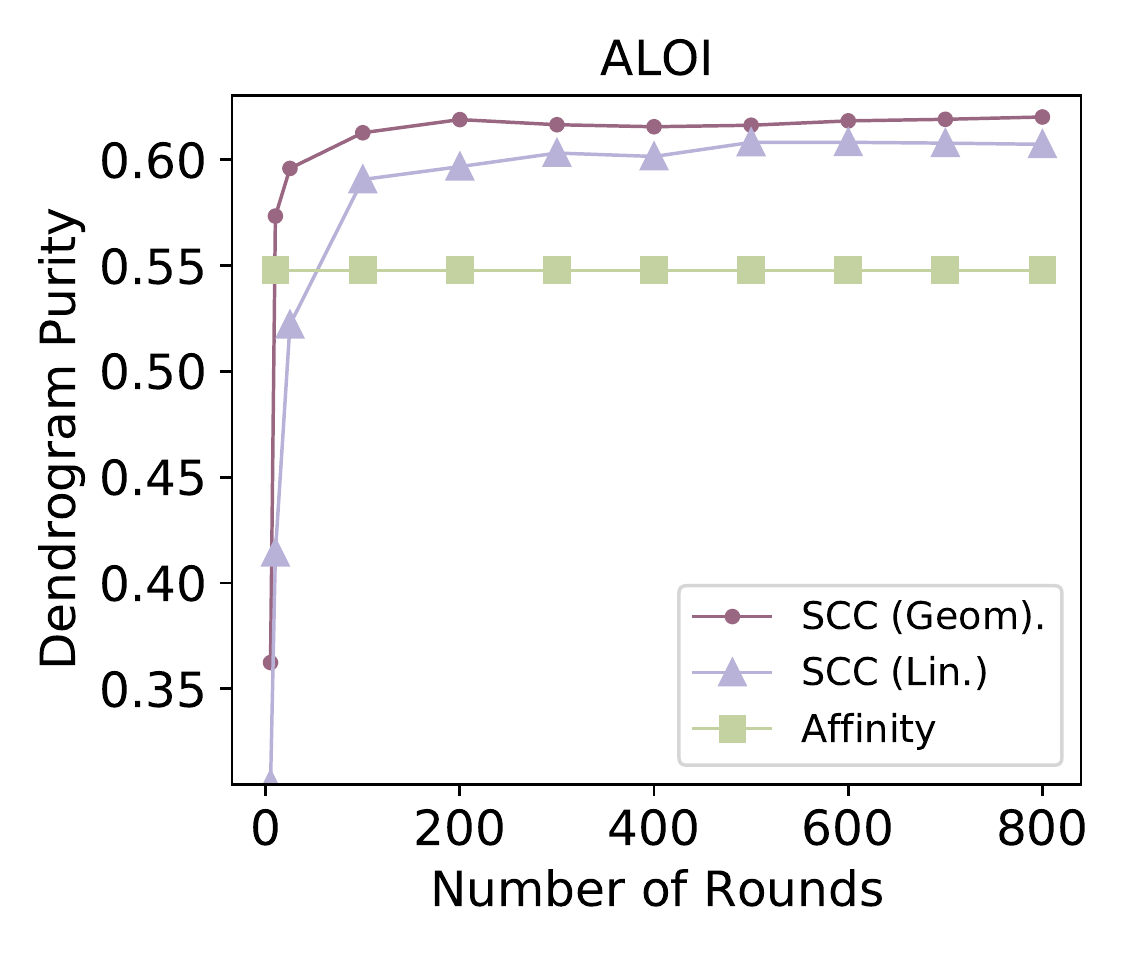}
    \includegraphics[width=0.4\textwidth]{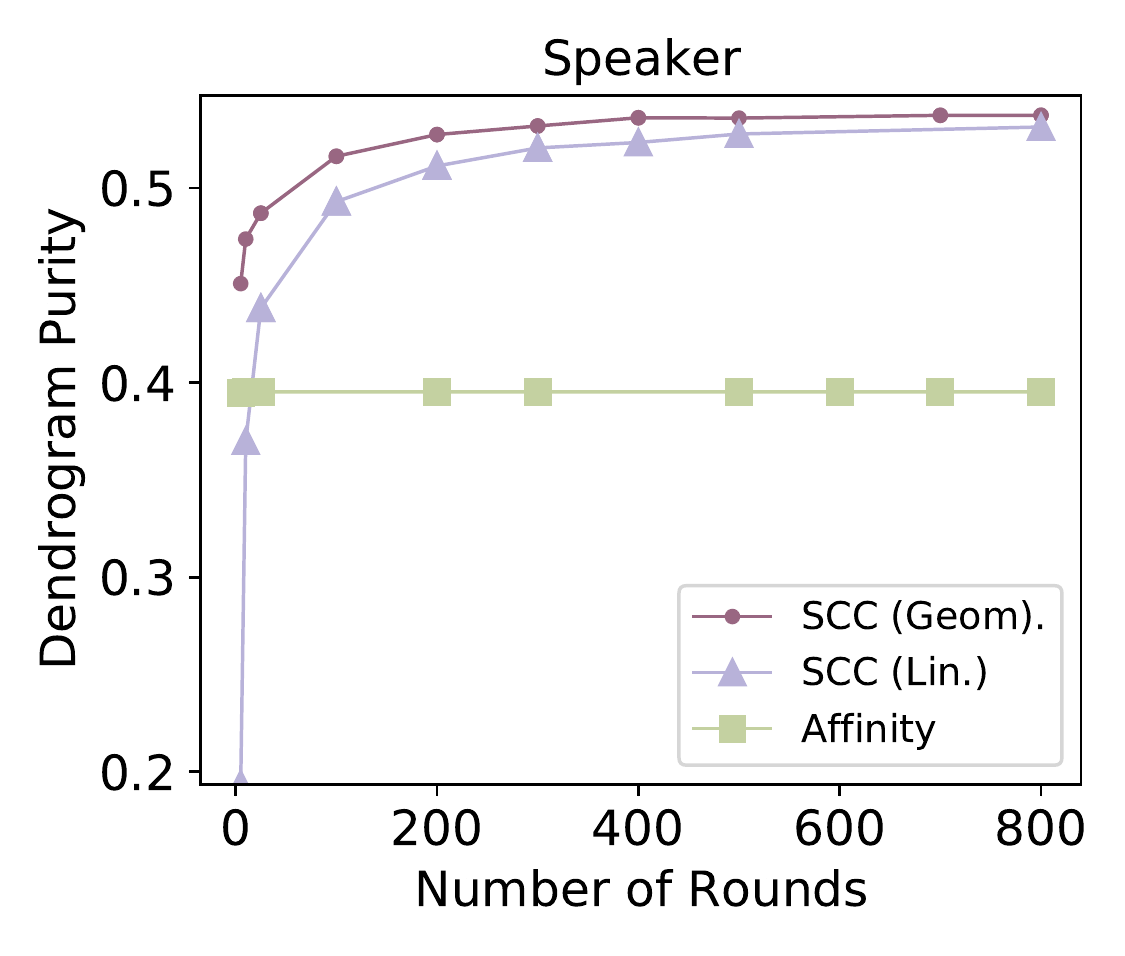}
    \vspace{-3mm}
    \caption{\textbf{Number of Round $\&$ Kinds of Thresholds}. 
    We report SCC's performance with both a geometric progression of thresholds and a linear progression of thresholds. We also report Affinity's performance, which converges in about five rounds. Recall that HAC would require number of rounds equal to number of points (much more than 800).}
    \label{fig:rounds}\vspace{-3mm}
\end{figure}

We analyze the decision to use geometric progression for
thresholds as compared to linear ones in Figure~\ref{fig:rounds}. 
We observe that \alg performs slightly better with geometric
sequences compared to linear ones. We also observe that 
\alg requires only a few more rounds than Affinity to begin to
achieve high quality results and that the performance saturates  
after a few hundred rounds.

\begin{figure}
    \centering
    \includegraphics[width=.44\textwidth]{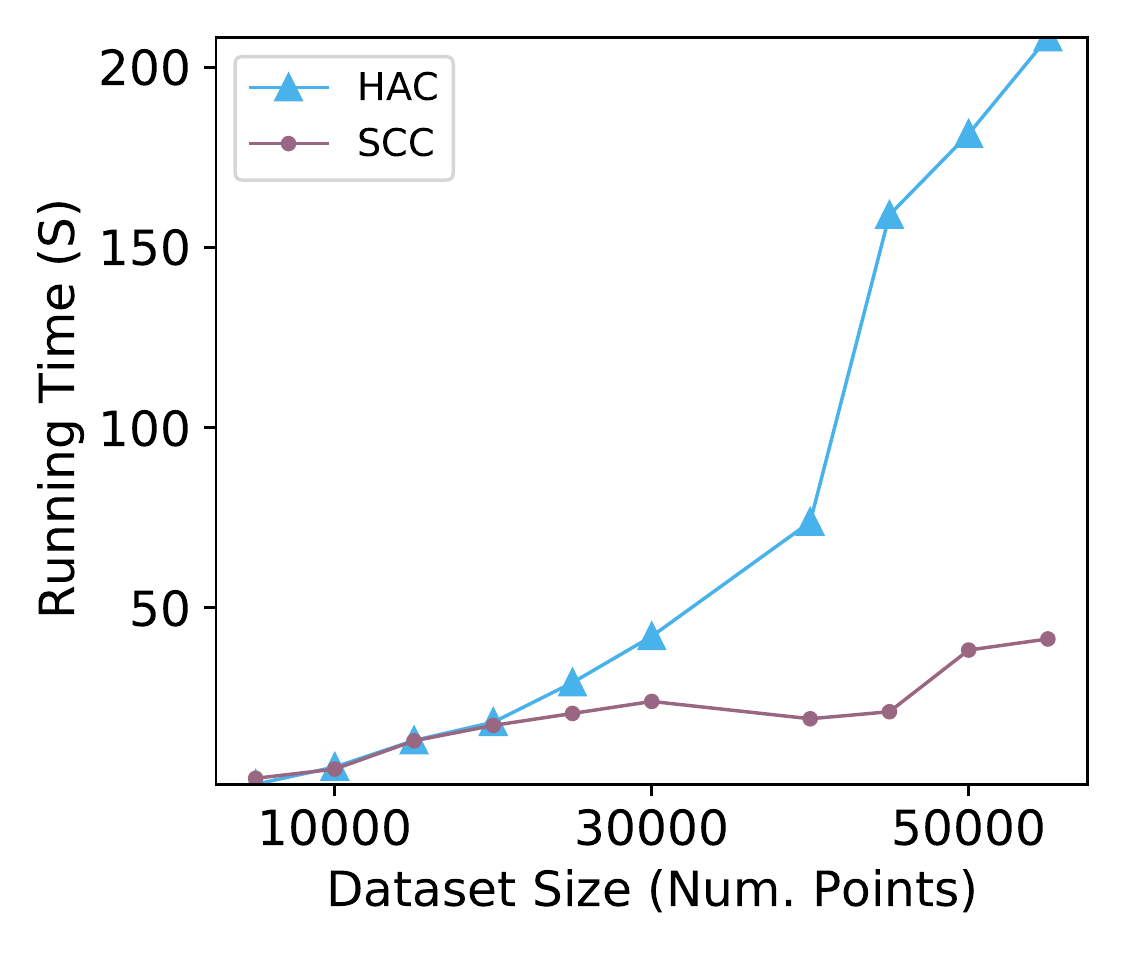}
    \includegraphics[width=.44\textwidth]{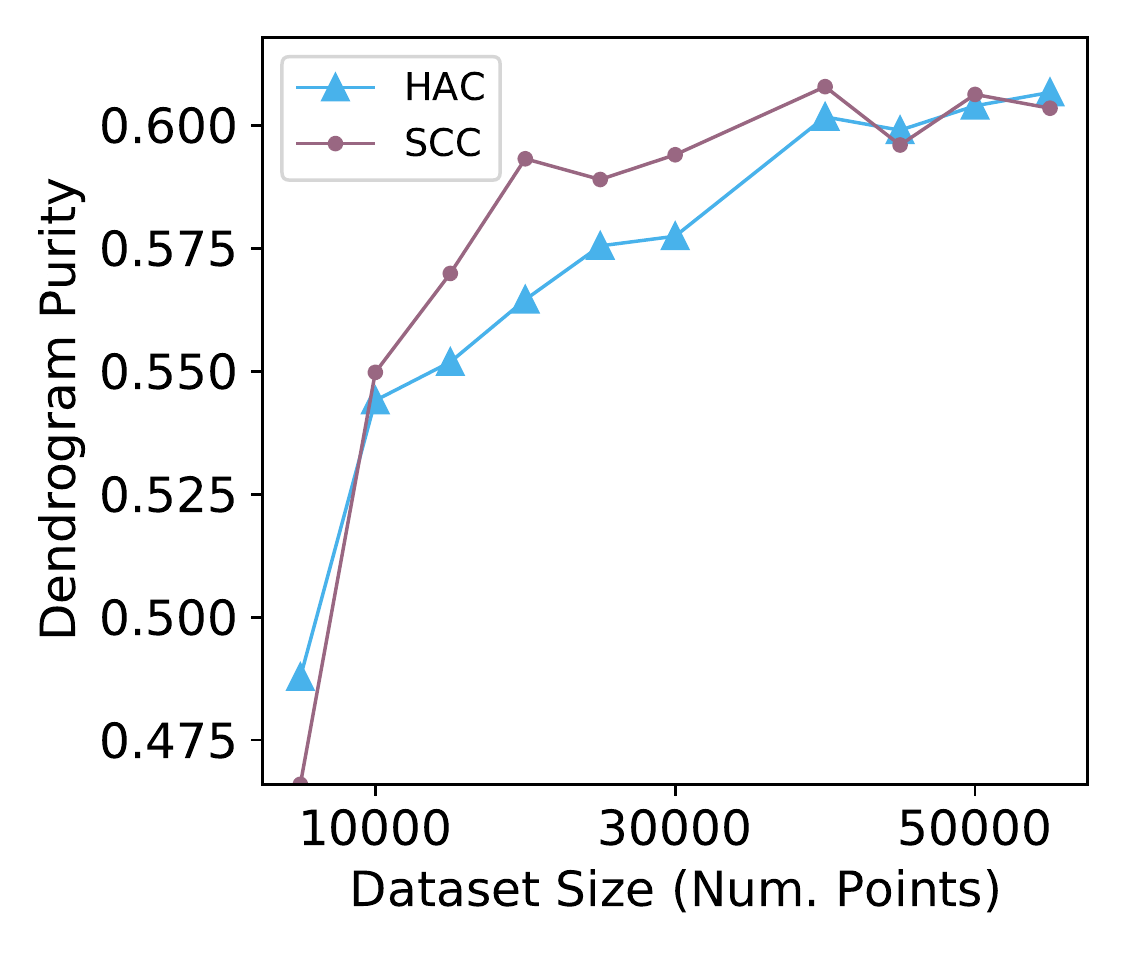}
        \caption{\textbf{Comparison to HAC}: We report running time and Dendrogram Purity of \scc compared to HAC, on a synthetic dataset. }
    \label{fig:synth_hac_comparison}
\end{figure}

\noindent\textbf{Comparison to HAC}
 We sample datasets of varying sizes
from a Dirichlet Process Mixture Model. 
 Figure \ref{fig:synth_hac_comparison}
reports the running time and
dendrogram purity as a function of the dataset size.
We report \scc results with 200 rounds.
We observe that while the time complexity
of HAC grows quadratically, \alg 
remains much more constant. Despite being much more efficient than 
HAC we observe that \alg produces trees with comparable dendrogram 
purity. 

\subsection{Flat Clustering\label{sec:expt_fc}}
\begin{table}[t]
\centering
\footnotesize
\begin{tabular}{@{}l@{}rrrrrr@{}}
    \toprule
    & \bf \small{CovType} & \bf  \small{ILSVRC} &  \bf \small{ALOI} & \bf  \small{Spkr.}  & \bf  \small{ILSVRC} \\
    &  & \bf  \small{(Sm.)} &   &   & \bf  \small{(Lg.)} \\
    \midrule
     \textsc{Perch} & 0.230 & 0.543 & 0.442  & 0.318  & 0.257  \\
     K-Means  &  0.245 & 0.605  & 0.408 &  0.322 & 0.562 \\
     Affinity & 0.536  & \bf 0.632  &   0.439 &  0.299  &  \bf 0.641  \\
     \alg & \bf 0.536 & 0.609 & \bf 0.567 & \bf 0.493 &  0.602 \\
     \midrule
      Affinity (best F1)  & \it 0.536 & 0.632 & 0.465 & 0.314 & 0.641 \\
     \alg (best F1) & \it 0.536 & \it 0.654 & \it 0.605 & \it 0.526  & \it 0.664 \\
     \bottomrule
\end{tabular}
\caption{\textbf{Pairwise F1} results on benchmark datasets.}
\label{tbl:f1_all}
\end{table}

In this section, we empirically evaluate \scc against 
state-of-the-art approaches in terms of pairwise F1 of flat clusters discovered.
We use the same experimental setting as in previous works \citep{kobren2017hierarchical}. 
In this experiment, we use the ground truth number of clusters when selecting a flat clustering. We select the round 
of \alg which produced the closest number of clusters to the ground truth and report the flat clustering performance 
of that round. We do the same for baseline methods. We evaluate the quality of the flat clusterings using the 
pairwise F1 metric \citep{kobren2017hierarchical,manning2008introduction}, for which 
precision is defined as $\textsf{Prec} = \frac{| \mathcal{P}^* \cap \hat{\mathcal{P}} |}{|\hat{\mathcal{P}}|}$, and recall as $\textsf{Rec} = \frac{| \mathcal{P}^* \cap \hat{\mathcal{P}} |}{|{\mathcal{P}^*}|}$,
where $\mathcal{P}^*$ is all pairs of points that are 
assigned to the same cluster according to the ground truth $\clustering^*$
and  $\hat{\mathcal{P}}$ is similarly defined for the predicted clustering 
$\hat{\clustering}$:
    $\mathcal{P}^* = \{(x_i,x_j) \ | \ \ x_i,x_j \in \dataset, \ \exists \ C^* \in \clustering^* \text{ s.t. } \{x_i,x_j\} \subseteq C^* \}$ and
    $\hat{\mathcal{P}} = \{(x_i,x_j) \ | \ x_i,x_j \in \dataset, \ \exists \ C \in \hat{\clustering} \text{ s.t. } \{x_i,x_j\} \subseteq C \}$

Table \ref{tbl:f1_all} gives the F1 performance for each method on each of the datasets used in the hierarchical clustering experiments. We observe that both \alg and Affinity outperform the previous state-of-the-art results reported by \cite{kobren2017hierarchical}.  
 SCC is the best performing method on Speaker and ALOI and is competitive with Affinity on the remaining datasets. 
We report the best F1 achieved in \emph{any} round of our algorithm and Affinity, which is the next best performing method. SCC's best F1 is better than Affinity's. 

\subsection{Approximation of DP-Means Objective\label{sec:expt_dp}}

Our analysis section (\S \ref{sec:dpmeans}) showed that \scc is a DP-Means approximation algorithm. In this section, we empirically evaluate these claims. 
We measure the quality of the clustering discovered
by our algorithm in terms of the DP-Means objective. 

We compare \scc to the following state-of-the-art algorithms for obtaining solutions to the DP-means objective: 
 \textbf{SerialDPMeans}, the classic  iterative optimization algorithm for DP-Means \citep{broderick2013mad,jiang2012small,kulis2012revisiting,pan2013optimistic}, in which data point is added to a cluster if it is within $\lambda$ of that cluster center and otherwise starts a new cluster;
\textbf{DPMeans$++$} \citep{bachem2015coresets} an initialization-only method which performs a K-Means$++$ \citep{arthur2006k} style sampling procedure. For each method, we record the assignment of points to clusters given by inference. We use this assignment of points to flat clusters to produce a DP-Means cost. We perform our analysis on the aforementioned clustering benchmarks.

Figure \ref{fig:dpmeansplots} shows the DP-Means objective as a function of the value of the parameter $\lambda$. Figure \ref{fig:pairwise_eval} shows the corresponding F1 performance for different values of $\lambda$ 
(0.001, 0.005, 0.01, 0.05, 0.1, 0.25, 0.5, 0.75, 1.0, 1.25, 1.5, 1.75, 2.0). We report the min/max/average performance over multiple runs of the SerialDPMeans and DPMeans$++$ algorithms with different random seeds.
Here, each method uses normalized $\ell_2^2$ distance as the dissimilarity measure. \alg uses thresholds $0.001$ to $4$ with a geometric progression. 

For each value of $\lambda$, \scc 
achieves the lowest DP-Means cost, which we hypothesize is due to SCC discovering optimal clustering independently of $\lambda$ via multiple alternative partitions in the tree. 
As each algorithm uses the value of $\lambda$ differently, the value of $\lambda$ that results in the best F1 score on the dataset could be quite different for each method and for each dataset.
If we consider the best F1 value achieved, \scc is usually one of the best performing methods. On CovType, SerialDPMeans performs best. but F1 does not seem meaningful on CovType as we observed the highest F1 value (0.536) with all points in a single cluster. See Appendix \S \ref{sec:app_experiments} for ILSVRC (Lg.) results and comparison to  LowrankALBCD~\citep{yen2016scalable}.

\section{Application: Large Scale Clustering of Web Queries}
\label{exp:user_query}

\begin{figure*}
    \centering
    \includegraphics[width=.24\textwidth]{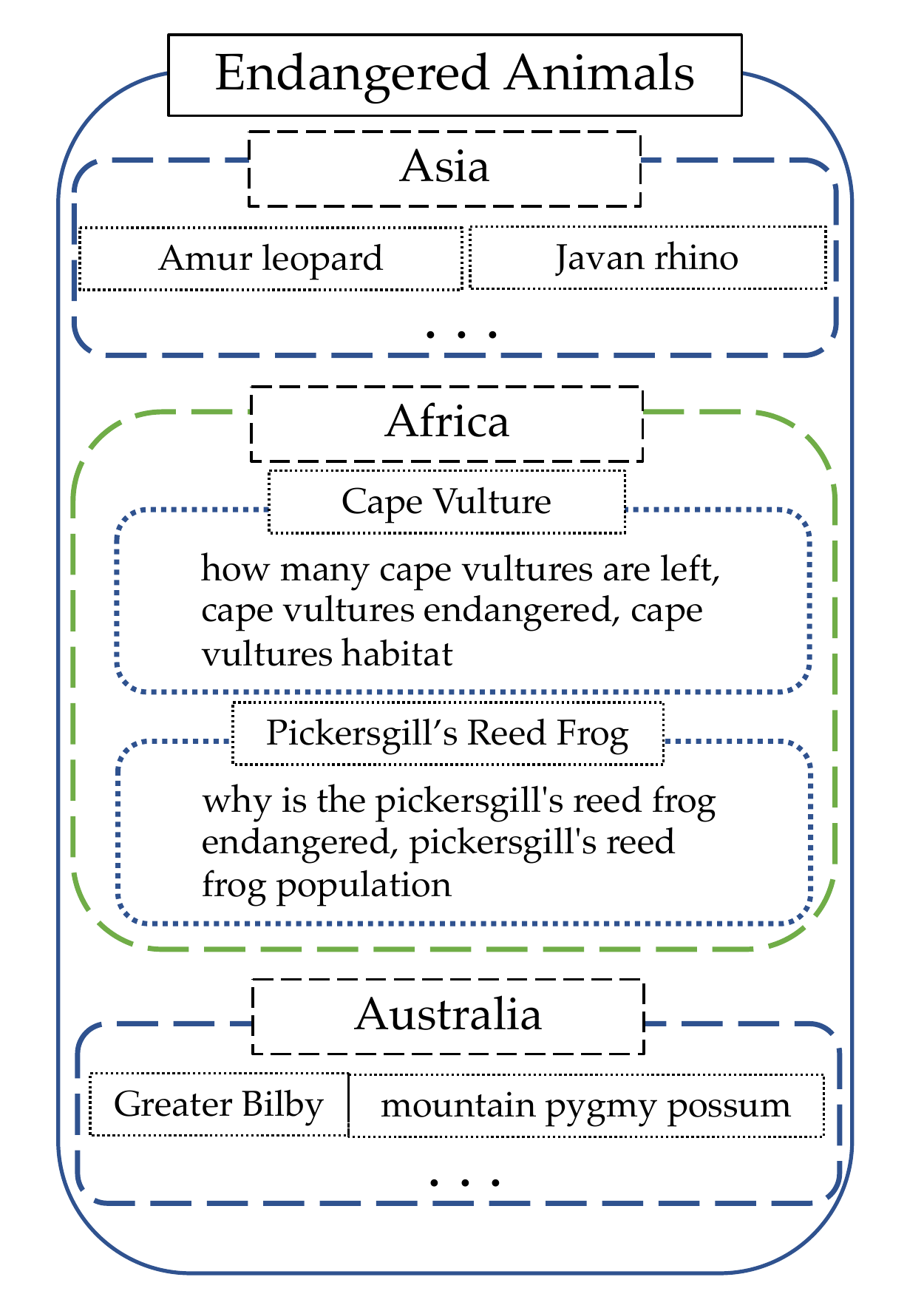}
    \includegraphics[width=.24\textwidth]{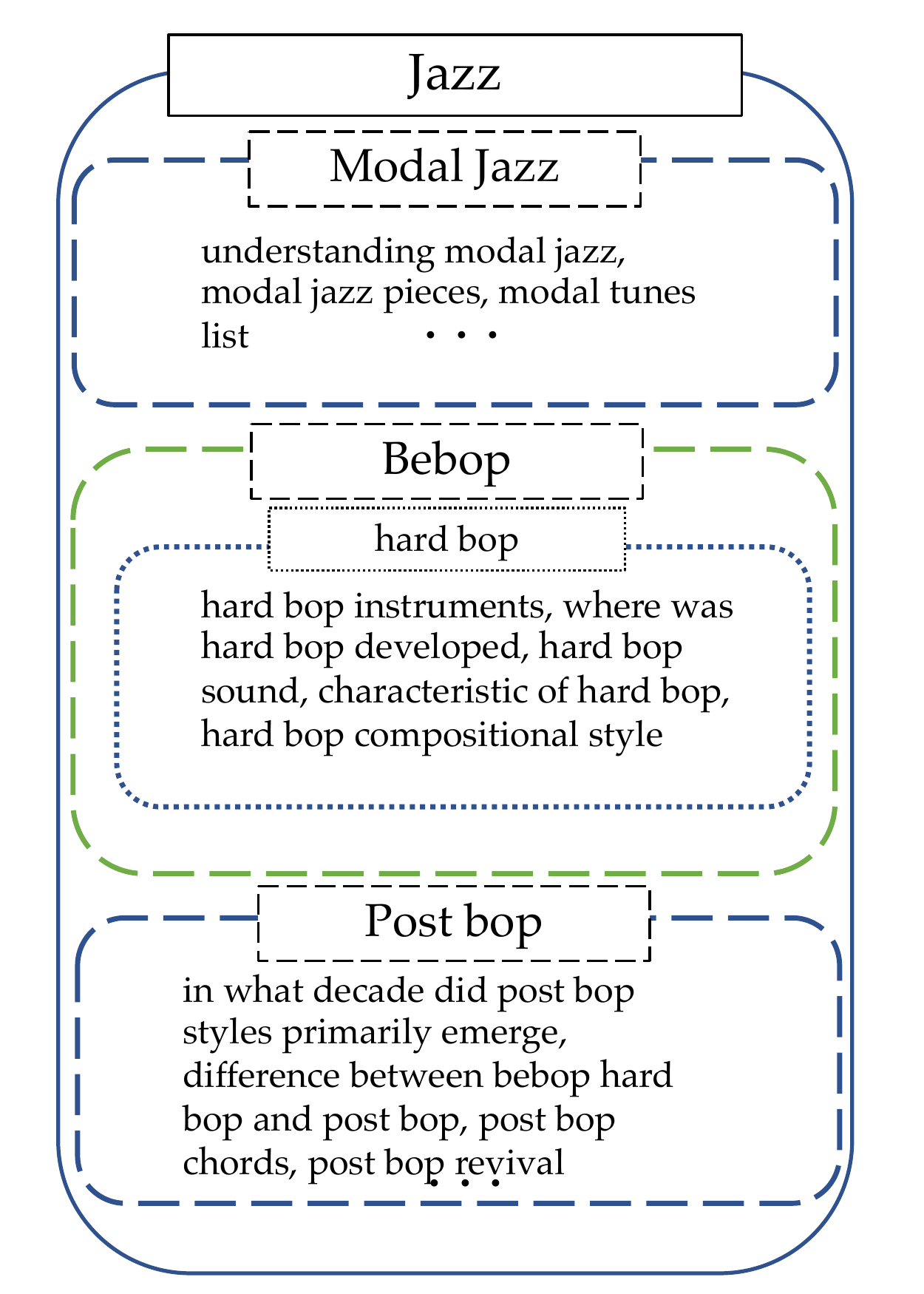}
    \includegraphics[width=.24\textwidth]{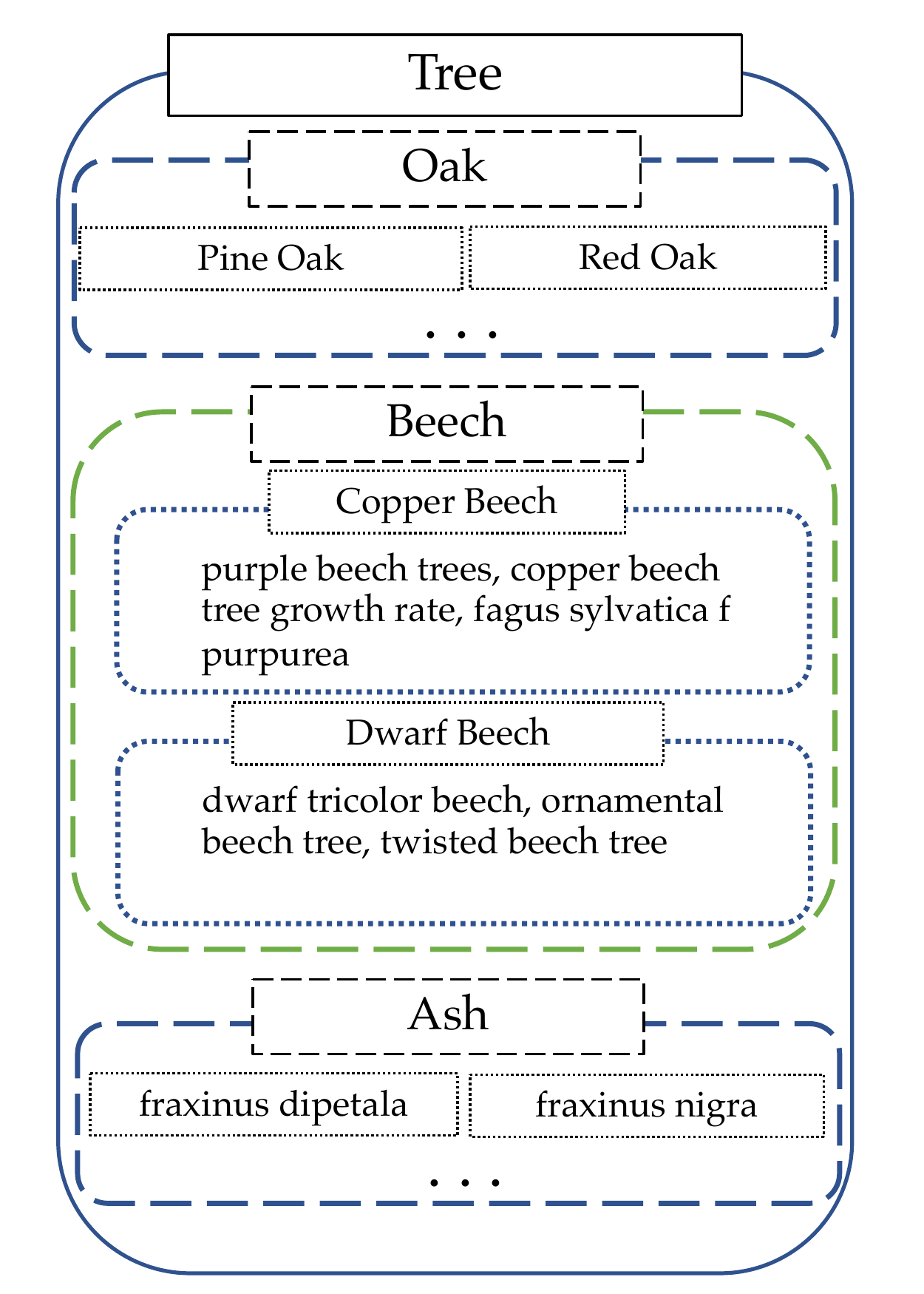}
    \includegraphics[width=.24\textwidth]{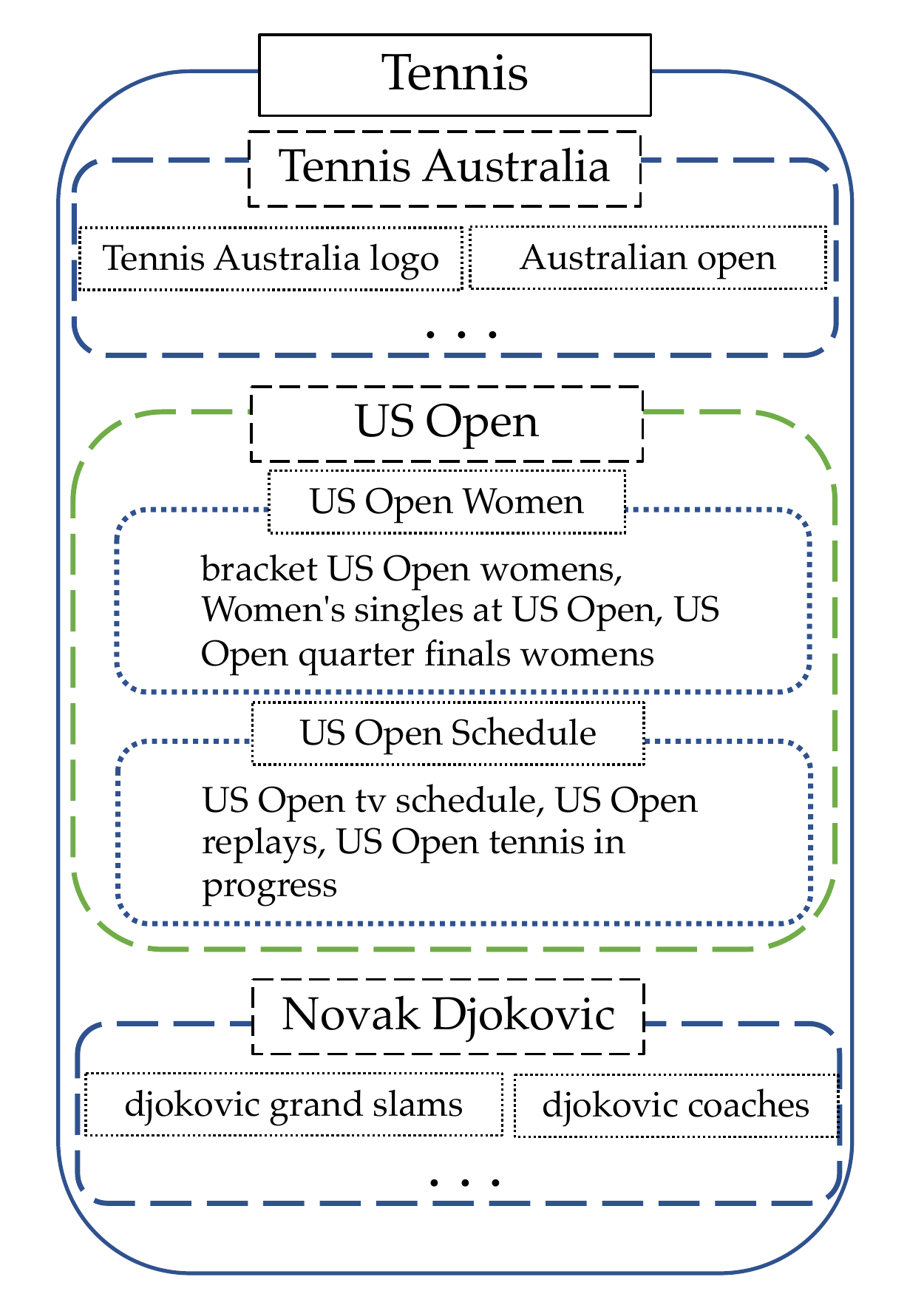}
    \caption{ Hierarchy inferred on 30 billion user queries using \alg. We represent the hierarchy using rectangular boxes. The root with header is represented by the outer rectangular box (solid line). The second level of the hierarchy, with header, is shown in dashed ($-\ -$) rectangle withing the outer box. Finally the third level from the root is shown as the inner most dotted ($...$) rectangle. Plain text within each of the dotted rectangle are the top queries that belong to that cluster. For example, \textsc{Endangered Animals}, is the root of the left most hierarchy, \textsc{Endangered Animals in Africa} is a sub-cluster and \textsc{pickergill's reed frog} is the lowest level cluster containing queries such as \textsc{why is the pickergill's reed frog endangered}.
    \label{fig:web_eg}}
\end{figure*}

We investigate the use of \alg for 
clustering web queries. We run our proposed clustering approach and Affinity clustering on a dataset of a random sample of \textbf{30 billion} queries. 
To the best of our knowledge this is one of the largest evaluations of any clustering algorithm. 
Due to the massive size of the data, we limit our evaluation to
the two most highly performing methods, \scc and Affinity clustering. 
Distance computation between queries during algorithm execution are sped-up using hashing techniques to avoid the $N^2$ pairwise dissimilarity bottleneck (for both \scc and Affinity).
Queries are represented using a set of proprietary features comprising lexical and behavioral signals among others. We extract manually a fine-grained level of flat clusterings and compared the clustering quality of the flat clusterings discovered by both algorithms. 
 
\begin{figure}
    \centering
     \includegraphics[width=.7\textwidth]{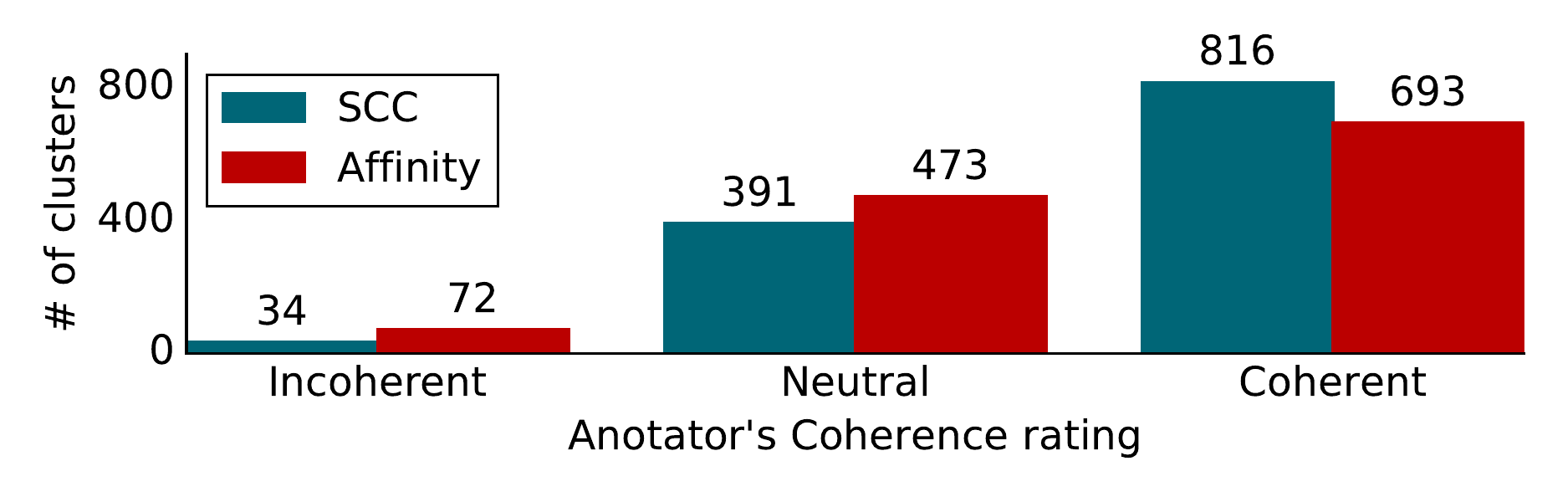}
    \caption{Human evaluation of clusters generated by \scc and Affinity \label{fig:human_eval}}
\end{figure}

 \textbf{Human Evaluation:} To evaluate the quality of flat clusters discovered by \alg, we conducted an empirical evaluation with human annotators. 
We asked them to rate $\sim1200$ randomly sampled  clusters from -1 (incoherent) to +1 (coherent). 
For example, a annotator might receive a head query of \texttt{home improvement} and a tail query of \texttt{lowes near me} from the same cluster.
The annotator then rates each of these pairs from -1 (incoherent) to +1 (coherent).
We report the aggregated results in Figure~\ref{fig:human_eval}.
We found that the annotators labeled 6.0\% of Affinity clustering's clusters and only 2.7\% of \scc clusters as incoherent. The annotators labeled 55.8\% of Affinity's clusters and 65.7\% of \scc clusters as coherent. 
We hope this demonstrates the cogency of the clusters found by SCC.

\textbf{Qualitative Evaluation:} 
In Table~\ref{tab:comparison_aff}, we show 
the clusters discovered by both SCC and Affinity that 
contain the query \emph{Green Velvet} (the house/techno music artist). We observe that SCC's cluster is considerably more precise and on topic.
 Table~\ref{tab:examples}, shows 
additional examples. We find the clusters to be precise    
and coherent. The algorithm discovers a clustering related to tennis strategies, which contains meaningful queries such as \emph{baseline tactics} and \emph{tennis strategy angles}. We  
investigate the hierarchical structure in Figure \ref{fig:web_eg}.
We discover meaningful clusters at multiple granularities with the tree structure indicating meaningful relationships between topics. For instance, we discover subgenres of \emph{jazz} such as \emph{bebop} and \emph{modal jazz}. We further discover that \emph{hard bop} 
is a subgenre of \emph{bebop}.

\begin{table}
    \centering
    \begin{tabular}{@{}lr@{}}
        \toprule
         \bf Affinity Clustering & \bf \alg \\ 
         \midrule
         green velvet live & green velvet mix \\
         green velvet talking & dj green velvet \\
         kestra financial businesswire & tomorrowland green velvet \\
         tomorrow land green velvet & green velvet 2015 \\
         dj green hair & green velvet music \\
         rinvelt \& david kestra businesswire &  green velvet dj set \\
         \bottomrule
    \end{tabular}
    \caption{\alg and Affinity Clustering for clusters corresponding to \emph{green velvet} on the 30 billion query dataset.}
    \label{tab:comparison_aff}
\end{table}

\begin{table}
\small
    \centering
    \begin{tabular}{@{}lcr@{}}
    \toprule
         \bf Tea Recipes  & \bf Electric Piano & \bf Tennis Strategy   \\
         \midrule
         tea drinks & digital piano price & playing strategies in tennis  \\
         tea recipes  & electric piano sale & understanding tennis tactics \\
         fancy tea recipes & electric piano small & baseline tactics. \\
         black tea flavors  & best digital piano ebay  & tennis strategy angles \\
         \bottomrule
    \end{tabular}
    \caption{Fine-grained Query Clusters Discovered by \alg}
    \label{tab:examples}
\end{table}

\section{Related Work}

Parallel and distributed approaches for hierarchical clustering
have been considered by previous work \citep{olson1995parallel,jin2015scalable,bateni2017affinity,yaroslavtsev2018massively,santos2019hierarchical}. 
\citet{yaroslavtsev2018massively} use a graph sparsification approach along with parallel minimum spanning tree approach to achieve provably good approximate minimum spanning trees.
Other work has proposed to achieve scalability by building hierarchical clusterings in an online, incremental, or streaming manner \citep{zhang1996birch,kranen2011clustree,kobren2017hierarchical,monath2019scalable}. BIRCH, one of the most widely used algorithms, builds a tree in a top down fashion splitting nodes under a condition on the mean/variance of the points assigned to a node. PERCH \citep{kobren2017hierarchical} and GRINCH \citep{monath2019scalable} add points next to the nearest neighbor node in the tree structure and perform tree re-arrangements.  \citet{kranen2011clustree} build an adaptive index structure for streaming data with allows for recency weighting.

Our work is closely related to the tree-based clustering methods proposed by \citet{balcan2008discriminative,balcan2014robust}. These methods also build a tree structure in a bottom up manner using round-specific thresholds to determine the mergers. These methods in fact recover clusterings under more flexible separation conditions than the one used in this paper. However, the linkage function computation used is more computationally expensive than the one used in this paper. We show an empirical comparison of this to \scc in Appendix \S \ref{sec:app_experiments} .

While this paper has focused on linkage-based hierarchical clustering in general, density-based clustering algorithms like DBSCAN \citep{ester1996density} correspond to specific linkage functions. There have been several hierarchical variants of these approaches proposed including, most recently HDBSCAN$^*$ \citep{campello2013density}. There are other related density and spanning-tree-based approaches \cite{cao2006density,zhu2019simple}.

Other work has use techniques to reduce the number of distance computations required to perform hierarchical
clustering \citep{eriksson2011active, krishnamurthy2012efficient}. \citet{krishnamurthy2012efficient} uses only $O(n\log^2(n))$ distance computations by repeatedly 
running a flat clustering algorithm on small subsets of the data to discover the children of each node in a top-down way. Other work uses nearest neighbor graph sparsification to reduce the complexity of algorithms \cite{zaheer2017canopy}. 

A variety of objective functions have been proposed for hierarchical clustering. Some work has used integer linear programming to perform hierarchical agglomerative clustering \citep{gilpin2013formalizing}. Notably, Dasgupta's cost function \citep{dasgupta2016cost} has been widely studied, including its connections to agglomerative clustering \citep{moseley2017approximation,charikar2019hierarchical}.

\section{Conclusion}
We introduce the Sub-Cluster Component
algorithm (\scc) for scalable 
hierarchical as well as flat clustering. 
\scc uses an agglomerative, round-based, approach 
in which a series of increasing distance thresholds
are used to determine which sub-clusters can 
be merged in a given round. 
We provide a theoretical analysis of \scc 
under well studied separability assumptions and relate
it to the non-parametric DP-means objective. 
We perform a comprehensive empirical analysis of \scc 
demonstrating its proficiency over state-of-the-art 
clustering algorithms on benchmark clustering datasets.
We further provide an analysis of the method with
respect to the DP-means objective. Finally, we demonstrate
the scalability of \scc by running on an industrial web-scale
dataset of 30B user queries. We evaluate clustering quality on this web-scale dataset with human annotations that indicate that the clusters produced by \scc are more coherent than state-of-the-art methods.

\section*{Acknowledgements}

We thank Avrim Blum and Nina Balcan for their insightful comments and suggestions. 
Andrew McCallum and Nicholas Monath are supported by the Center for Data Science and the Center for Intelligent Information
Retrieval and by the National Science Foundation under Grants No. 1763618.
Some of the work reported here was performed using high performance computing equipment obtained under a grant from the Collaborative R\&D Fund managed by the Massachusetts Technology Collaborative. Any opinions, findings and conclusions or recommendations expressed in this
material are those of the authors and do not necessarily reflect those of the sponsor.

\bibliographystyle{abbrvnat}
\bibliography{references}

\begin{thebibliography}{62}
\providecommand{\natexlab}[1]{#1}
\providecommand{\url}[1]{\texttt{#1}}
\expandafter\ifx\csname urlstyle\endcsname\relax
  \providecommand{\doi}[1]{doi: #1}\else
  \providecommand{\doi}{doi: \begingroup \urlstyle{rm}\Url}\fi

\bibitem[Andrews et~al.(2014)Andrews, Eisner, and Dredze]{andrews2014robust}
N.~Andrews, J.~Eisner, and M.~Dredze.
\newblock Robust entity clustering via phylogenetic inference.
\newblock \emph{ACL}, 2014.

\bibitem[Arthur and Vassilvitskii(2007)]{arthur2006k}
D.~Arthur and S.~Vassilvitskii.
\newblock k-means++: The advantages of careful seeding.
\newblock \emph{SODA}, 2007.

\bibitem[Bachem et~al.(2015)Bachem, Lucic, and Krause]{bachem2015coresets}
O.~Bachem, M.~Lucic, and A.~Krause.
\newblock Coresets for nonparametric estimation-the case of dp-means.
\newblock \emph{ICML}, 2015.

\bibitem[Balcan et~al.(2008)Balcan, Blum, and
  Vempala]{balcan2008discriminative}
M.~Balcan, A.~Blum, and S.~Vempala.
\newblock A discriminative framework for clustering via similarity functions.
\newblock \emph{STOC}, 2008.

\bibitem[Balcan et~al.(2014)Balcan, Liang, and Gupta]{balcan2014robust}
M.~Balcan, Y.~Liang, and P.~Gupta.
\newblock Robust hierarchical clustering.
\newblock \emph{JMLR}, 2014.

\bibitem[Bateni et~al.(2017)Bateni, Behnezhad, Derakhshan, Hajiaghayi, Kiveris,
  Lattanzi, and Mirrokni]{bateni2017affinity}
M.~Bateni, S.~Behnezhad, M.~Derakhshan, M.~Hajiaghayi, R.~Kiveris, S.~Lattanzi,
  and V.~Mirrokni.
\newblock Affinity clustering: Hierarchical clustering at scale.
\newblock \emph{NeurIPS}, 2017.

\bibitem[Benevenuto et~al.(2012)Benevenuto, Rodrigues, Cha, and
  Almeida]{Benevenuto2012-ml}
F.~Benevenuto, T.~Rodrigues, M.~Cha, and V.~Almeida.
\newblock Characterizing user navigation and interactions in online social
  networks.
\newblock \emph{Inf. Sci.}, 2012.

\bibitem[Beygelzimer et~al.(2006)Beygelzimer, Kakade, and
  Langford]{beygelzimer2006cover}
A.~Beygelzimer, S.~Kakade, and J.~Langford.
\newblock Cover trees for nearest neighbor.
\newblock \emph{ICML}, 2006.

\bibitem[Blundell et~al.(2010)Blundell, Teh, and Heller]{blundell2010bayesian}
C.~Blundell, Y.~W. Teh, and K.~A. Heller.
\newblock Bayesian rose trees.
\newblock \emph{UAI}, 2010.

\bibitem[Bor{\r{u}}vka(1926)]{boruuvka1926jistem}
O.~Bor{\r{u}}vka.
\newblock O jist{\'e}m probl{\'e}mu minim{\'a}ln{\'\i}m.
\newblock 1926.

\bibitem[Broderick et~al.(2013)Broderick, Kulis, and Jordan]{broderick2013mad}
T.~Broderick, B.~Kulis, and M.~Jordan.
\newblock Mad-bayes: Map-based asymptotic derivations from bayes.
\newblock \emph{ICML}, 2013.

\bibitem[Campello et~al.(2013)Campello, Moulavi, and
  Sander]{campello2013density}
R.~J. G.~B. Campello, D.~Moulavi, and J.~Sander.
\newblock Density-based clustering based on hierarchical density estimates.
\newblock \emph{Advances in Knowledge Discovery and Data Mining}, 2013.

\bibitem[Cao et~al.(2006)Cao, Ester, Qian, and Zhou]{cao2006density}
F.~Cao, M.~Ester, W.~Qian, and A.~Zhou.
\newblock Density-based clustering over an evolving data stream with noise.
\newblock \emph{ICDM}, 2006.

\bibitem[Charikar et~al.(2019)Charikar, Chatziafratis, and
  Niazadeh]{charikar2019hierarchical}
M.~Charikar, V.~Chatziafratis, and R.~Niazadeh.
\newblock Hierarchical clustering better than average-linkage.
\newblock \emph{SODA}, 2019.

\bibitem[Chen et~al.(2018)Chen, Shrivastava, Steorts, et~al.]{chen2018unique}
B.~Chen, A.~Shrivastava, R.~C. Steorts, et~al.
\newblock Unique entity estimation with application to the syrian conflict.
\newblock \emph{The Annals of Applied Statistics}, 2018.

\bibitem[Cranmer et~al.(2019)Cranmer, Macaluso, and
  Pappadopulo]{cranmer2019toy}
K.~Cranmer, S.~Macaluso, and D.~Pappadopulo.
\newblock Toy generative model for jets.
\newblock 2019.

\bibitem[Das et~al.(2020)Das, Godbole, Monath, Zaheer, and
  Mc{C}allum]{das2020probabilistic}
R.~Das, A.~Godbole, N.~Monath, M.~Zaheer, and A.~Mc{C}allum.
\newblock Probabilistic case-based reasoning for open-world knowledge graph
  completion.
\newblock \emph{Findings of EMNLP}, 2020.

\bibitem[Dasgupta(2016)]{dasgupta2016cost}
S.~Dasgupta.
\newblock A cost function for similarity-based hierarchical clustering.
\newblock \emph{Symposium on Theory of Computing (STOC)}, 2016.

\bibitem[Ein~Dor et~al.()Ein~Dor, Mass, Halfon, Venezian, Shnayderman,
  Aharonov, and Slonim]{ein-dor-etal-2018-learning}
L.~Ein~Dor, Y.~Mass, A.~Halfon, E.~Venezian, I.~Shnayderman, R.~Aharonov, and
  N.~Slonim.
\newblock Learning thematic similarity metric from article sections using
  triplet networks.
\newblock \emph{ACL}.

\bibitem[Eisen et~al.(1998)Eisen, Spellman, Brown, and
  Botstein]{eisen1998cluster}
M.~B. Eisen, P.~T. Spellman, P.~O. Brown, and D.~Botstein.
\newblock Cluster analysis and display of genome-wide expression patterns.
\newblock \emph{Proceedings of the National Academy of Sciences}, 1998.

\bibitem[Eriksson et~al.(2011)Eriksson, Dasarathy, Singh, and
  Nowak]{eriksson2011active}
B.~Eriksson, G.~Dasarathy, A.~Singh, and R.~Nowak.
\newblock Active clustering: Robust and efficient hierarchical clustering using
  adaptively selected similarities.
\newblock \emph{AISTATS}, 2011.

\bibitem[Ester et~al.(1996)Ester, Kriegel, Sander, Xu,
  et~al.]{ester1996density}
M.~Ester, H.-P. Kriegel, J.~Sander, X.~Xu, et~al.
\newblock A density-based algorithm for discovering clusters in large spatial
  databases with noise.
\newblock \emph{KDD}, 1996.

\bibitem[Gavryushkin and Drummond(2016)]{gavryushkin2016space}
A.~Gavryushkin and A.~J. Drummond.
\newblock The space of ultrametric phylogenetic trees.
\newblock \emph{Journal of theoretical biology}, 2016.

\bibitem[Geusebroek et~al.(2005)Geusebroek, Burghouts, and
  Smeulders]{geusebroek2005amsterdam}
J.~Geusebroek, G.~J. Burghouts, and A.~W.~M. Smeulders.
\newblock The amsterdam library of object images.
\newblock \emph{IJCV}, 2005.

\bibitem[Gilpin et~al.(2013)Gilpin, Nijssen, and
  Davidson]{gilpin2013formalizing}
S.~Gilpin, S.~Nijssen, and I.~Davidson.
\newblock Formalizing hierarchical clustering as integer linear programming.
\newblock \emph{AAAI}, 2013.

\bibitem[Green et~al.(2012)Green, Andrews, Gormley, Dredze, and
  Manning]{green2012entity}
S.~Green, N.~Andrews, M.~R. Gormley, M.~Dredze, and C.~D. Manning.
\newblock Entity clustering across languages.
\newblock \emph{NAACL-HLT}, 2012.

\bibitem[Greenberg~et al(2014)]{greenberg2014nist}
C.~S. Greenberg~et al.
\newblock The nist 2014 speaker recognition i-vector machine learning
  challenge.
\newblock \emph{Odyssey: The Speaker and Language Recognition Workshop}, 2014.

\bibitem[Gro{\ss}wendt et~al.(2019)Gro{\ss}wendt, R{\"o}glin, and
  Schmidt]{grosswendt2019analysis}
A.~Gro{\ss}wendt, H.~R{\"o}glin, and M.~Schmidt.
\newblock Analysis of ward's method.
\newblock \emph{SODA}, 2019.

\bibitem[Guo et~al.(2020)Guo, Sun, Lindgren, Geng, Simcha, Chern, , and
  Kumar]{avq_2020}
R.~Guo, P.~Sun, E.~Lindgren, Q.~Geng, D.~Simcha, F.~Chern, , and S.~Kumar.
\newblock Accelerating large-scale inference with anisotropic vector
  quantization.
\newblock In \emph{ICML}, 2020.

\bibitem[Heller and Ghahramani(2005{\natexlab{a}})]{heller2005randomized}
K.~Heller and Z.~Ghahramani.
\newblock Randomized algorithms for fast bayesian hierarchical clustering.
\newblock \emph{EU-PASCAL Statistics and Optimization of Clustering Workshop},
  2005{\natexlab{a}}.

\bibitem[Heller and Ghahramani(2005{\natexlab{b}})]{heller2005bayesian}
K.~A. Heller and Z.~Ghahramani.
\newblock Bayesian hierarchical clustering.
\newblock \emph{ICML}, 2005{\natexlab{b}}.

\bibitem[Jiang et~al.(2012)Jiang, Kulis, and Jordan]{jiang2012small}
K.~Jiang, B.~Kulis, and M.~I. Jordan.
\newblock Small-variance asymptotics for exponential family dirichlet process
  mixture models.
\newblock \emph{NeurIPS}, 2012.

\bibitem[Jin~et al(2015)]{jin2015scalable}
C.~Jin~et al.
\newblock A scalable hierarchical clustering algorithm using spark.
\newblock \emph{BigDataService}, 2015.

\bibitem[Kenyon-Dean et~al.(2018)Kenyon-Dean, Cheung, and
  Precup]{dean2018resolving}
K.~Kenyon-Dean, J.~C.~K. Cheung, and D.~Precup.
\newblock Resolving event coreference with supervised representation learning
  and clustering-oriented regularization.
\newblock \emph{*SEM}, 2018.

\bibitem[Kleinberg(2002)]{kleinberg2002impossibility}
J.~Kleinberg.
\newblock An impossibility theorem for clustering.
\newblock \emph{NeurIPS}, 2002.

\bibitem[Kobren et~al.(2017)Kobren, Monath, Krishnamurthy, and
  Mc{C}allum]{kobren2017hierarchical}
A.~Kobren, N.~Monath, A.~Krishnamurthy, and A.~Mc{C}allum.
\newblock A hierarchical algorithm for extreme clustering.
\newblock \emph{KDD}, 2017.

\bibitem[Kobren et~al.(2019)Kobren, Monath, and
  Mc{C}allum]{kobren2019integrating}
A.~Kobren, N.~Monath, and A.~Mc{C}allum.
\newblock Integrating user feedback under identity uncertainty in knowledge
  base construction.
\newblock \emph{AKBC}, 2019.

\bibitem[Kranen et~al.(2011)Kranen, Assent, Baldauf, and
  Seidl]{kranen2011clustree}
P.~Kranen, I.~Assent, C.~Baldauf, and T.~Seidl.
\newblock The clustree: indexing micro-clusters for anytime stream mining.
\newblock \emph{KIS}, 2011.

\bibitem[Krishnamurthy et~al.(2012)Krishnamurthy, Balakrishnan, Xu, and
  Singh]{krishnamurthy2012efficient}
A.~Krishnamurthy, S.~Balakrishnan, M.~Xu, and A.~Singh.
\newblock Efficient active algorithms for hierarchical clustering.
\newblock \emph{ICML}, 2012.

\bibitem[Kulis and Jordan(2012)]{kulis2012revisiting}
B.~Kulis and M.~I. Jordan.
\newblock Revisiting k-means: New algorithms via bayesian nonparametrics.
\newblock \emph{ICML}, 2012.

\bibitem[Kushagra et~al.(2016)Kushagra, Samadi, and Ben-David]{KSB16}
S.~Kushagra, S.~Samadi, and S.~Ben-David.
\newblock Finding meaningful cluster structure amidst background noise.
\newblock \emph{ALT}, 2016.

\bibitem[Lee et~al.(2012)Lee, Recasens, Chang, Surdeanu, and
  Jurafsky]{lee2012joint}
H.~Lee, M.~Recasens, A.~Chang, M.~Surdeanu, and D.~Jurafsky.
\newblock Joint entity and event coreference resolution across documents.
\newblock \emph{EMNLP-CoNLL}, 2012.

\bibitem[Mahajan et~al.(2009)Mahajan, Nimbhorkar, and
  Varadarajan]{mahajan2009planar}
M.~Mahajan, P.~Nimbhorkar, and K.~Varadarajan.
\newblock The planar k-means problem is np-hard.
\newblock \emph{WALCOM}, 2009.

\bibitem[Manning et~al.(2008)Manning, Raghavan, and
  Sch\"{u}tze]{manning2008introduction}
C.~D. Manning, P.~Raghavan, and H.~Sch\"{u}tze.
\newblock \emph{Introduction to Information Retrieval}.
\newblock 2008.

\bibitem[Monath et~al.(2019{\natexlab{a}})Monath, Kobren, Krishnamurthy, Glass,
  and Mc{C}allum]{monath2019scalable}
N.~Monath, A.~Kobren, A.~Krishnamurthy, M.~R. Glass, and A.~Mc{C}allum.
\newblock Scalable hierarchical clustering with tree grafting.
\newblock \emph{KDD}, 2019{\natexlab{a}}.

\bibitem[Monath et~al.(2019{\natexlab{b}})Monath, Zaheer, Silva, Mc{C}allum,
  and Ahmed]{monath2019gradient}
N.~Monath, M.~Zaheer, D.~Silva, A.~Mc{C}allum, and A.~Ahmed.
\newblock Gradient-based hierarchical clustering using continuous
  representations of trees in hyperbolic space.
\newblock \emph{KDD}, 2019{\natexlab{b}}.

\bibitem[Moseley and Wang(2017)]{moseley2017approximation}
B.~Moseley and J.~Wang.
\newblock Approximation bounds for hierarchical clustering: Average linkage,
  bisecting k-means, and local search.
\newblock \emph{NeurIPS}, 2017.

\bibitem[Olson(1995)]{olson1995parallel}
C.~F. Olson.
\newblock Parallel algorithms for hierarchical clustering.
\newblock \emph{Parallel computing}, 1995.

\bibitem[Pan et~al.(2013)Pan, Gonzalez, Jegelka, Broderick, and
  Jordan]{pan2013optimistic}
X.~Pan, J.~E. Gonzalez, S.~Jegelka, T.~Broderick, and M.~I. Jordan.
\newblock Optimistic concurrency control for distributed unsupervised learning.
\newblock \emph{NeurIPS}, 2013.

\bibitem[Russakovsky~et al(2015)]{russakovsky2015imagenet}
O.~Russakovsky~et al.
\newblock Imagenet large scale visual recognition challenge.
\newblock \emph{IJCV}, 2015.

\bibitem[Santos et~al.(2019)Santos, Syed, Naldi, Campello, and
  Sander]{santos2019hierarchical}
J.~Santos, T.~Syed, M.~C. Naldi, R.~J. Campello, and J.~Sander.
\newblock Hierarchical density-based clustering using mapreduce.
\newblock \emph{TBD}, 2019.

\bibitem[Schwartz~et al(2020)]{Schwartz2020-da}
G.~W. Schwartz~et al.
\newblock {TooManyCells} identifies and visualizes relationships of single-cell
  clades.
\newblock \emph{Nat. Methods}, 2020.

\bibitem[Vazirani(2013)]{vazirani2013approximation}
V.~V. Vazirani.
\newblock \emph{Approximation algorithms}.
\newblock 2013.

\bibitem[Vitale et~al.(2019)Vitale, Rajagopalan, and
  Gentile]{vitale2019flattening}
F.~Vitale, A.~Rajagopalan, and C.~Gentile.
\newblock Flattening a hierarchical clustering through active learning.
\newblock \emph{NeurIPS}, 2019.

\bibitem[Yadav et~al.(2019)Yadav, Kobren, Monath, and
  Mc{C}allum]{pmlr-v97-yadav19a}
N.~Yadav, A.~Kobren, N.~Monath, and A.~Mc{C}allum.
\newblock Supervised hierarchical clustering with exponential linkage.
\newblock \emph{ICML}, 2019.

\bibitem[Yaroslavtsev and Vadapalli(2018)]{yaroslavtsev2018massively}
G.~Yaroslavtsev and A.~Vadapalli.
\newblock Massively parallel algorithms and hardness for single-linkage
  clustering under $\ell_p$ distances.
\newblock \emph{ICML}, 2018.

\bibitem[Yen et~al.(2016)Yen, Malioutov, and Kumar]{yen2016scalable}
I.~E. Yen, D.~Malioutov, and A.~Kumar.
\newblock Scalable exemplar clustering and facility location via augmented
  block coordinate descent with column generation.
\newblock \emph{AISTATS}, 2016.

\bibitem[Zaheer et~al.(2017)Zaheer, Kottur, Ahmed, Moura, and
  Smola]{zaheer2017canopy}
M.~Zaheer, S.~Kottur, A.~Ahmed, J.~Moura, and A.~Smola.
\newblock Canopy---fast sampling with cover trees.
\newblock \emph{ICML}, 2017.

\bibitem[Zhang et~al.(1996)Zhang, Ramakrishnan, and Livny]{zhang1996birch}
T.~Zhang, R.~Ramakrishnan, and M.~Livny.
\newblock Birch: an efficient data clustering method for very large databases.
\newblock \emph{SIGMOD}, 1996.

\bibitem[Zhang et~al.(2014)Zhang, Ahmed, Josifovski, and
  Smola]{zhang2014taxonomy}
Y.~Zhang, A.~Ahmed, V.~Josifovski, and A.~Smola.
\newblock Taxonomy discovery for personalized recommendation.
\newblock \emph{WSDM}, 2014.

\bibitem[Zhang et~al.(2018)Zhang, Zhang, Yao, and Tang]{Zhang2018-pf}
Y.~Zhang, F.~Zhang, P.~Yao, and J.~Tang.
\newblock Name disambiguation in {AMiner}: Clustering, maintenance, and human
  in the loop.
\newblock \emph{KDD}, 2018.

\bibitem[Zhu and Stuetzle(2019)]{zhu2019simple}
H.~Zhu and W.~Stuetzle.
\newblock A simple and efficient method to compute a single linkage dendrogram.
\newblock \emph{arXiv}, 2019.

\end{thebibliography}

\clearpage
\appendix
{
\begin{center}
    \Large
    \textbf{{ Scalable Hierarchical Agglomerative Clustering -- Appendix}}
\end{center}
}

\section{Proofs}

\subsection{Proof of Theorem \ref{thm:opt}}
\label{proof:thm2}
Here we show a proof for the $\ell_2^2$ case, the $\ell_2$ case follows the same proof but uses the triangle inequality instead of the relaxed triangle inequality.
Recall the assumption of $\delta$-separability (Assumption \ref{asmp:delta-sep}) in which the maximum distance from any point to its 
true center is defined as $R:= \max_{i \in [k] } \max_{x \in C^*_i} \twoNormSqr{x - C^*_i}$. We make the additional assumption that the threshold of the first round, $\tau_0$, is less than $R$, i.e., $\tau_0 < R$.

The algorithm begins with $\clustering^{(0)}$ set to be the shattered partition, with each data point in its own cluster, $\clustering^{(0)} = \{\{x\}|x \in \dataset\}$. 

We want to show that some round, $r^\star$ with threshold $\tau_r$
produces the ground truth partition, $\clustering^{(r^\star)} = \clustering^* = \{C_1^*, \dots, C_k^*\}$. We will show by induction that for each round prior $r' \leq r^\star$ that the clustering $\clustering^{(r')}$ is $\emph{pure}$, i.e. $\forall C \in \clustering^{(r')}, \ \exists C^* \in \clustering^* \ C \subseteq C^*$ (equality will be for round $r^\star$). We will show  the round $r^\star$  with $\clustering^{(r^\star)} = \clustering^*$ must exist.

In our inductive hypothesis, we assume that for rounds before and including $\tau_r$, we have \emph{pure} sub-clusters such that $X, X' \subseteq C^*_i$, which are disjoint, $X \cap X' = \emptyset$, and $Y \subseteq C^*_j$ for
 $i \neq j$ ($\clustering^{(0)}$ by definition has pure sub-clusters). We want to show: every such $X$ and $X'$ must form a sub-cluster component without any such $Y$. In this way, we ensure that $C^*_i$ exists as a pure cluster in some round.
Using the relaxed triangle inequality \citep{grosswendt2019analysis} for $\ell_2^2$, we have:
{\tiny \begin{align*}\vspace{-1mm}
\twoNormSqr{c_i^* - c_j^*} & \leq 3 \left ( \frac{1}{|X||Y|} \sum_{x \in X}\sum_{y \in Y} \twoNormSqr{c_i^* - x} + \twoNormSqr{x - y} +  \twoNormSqr{y - c_j^*} \right )\\
\twoNormSqr{c_i^* - c_j^*} &\leq 3 \left ( \frac{1}{|X|} \sum_{x \in X} \twoNormSqr{c_i^* - x} + \frac{1}{|X||Y|} \sum_{x \in X}\sum_{y \in Y} \twoNormSqr{x - y} + \frac{1}{|Y|} \sum_{y \in Y} \twoNormSqr{y - c_j^*} \right )
\end{align*}}\vspace{-2mm}
{\tiny \begin{align*}\vspace{-2mm}
 \frac{1}{3}\twoNormSqr{c_i^* - c_j^*} - \frac{1}{|X|}  \sum_{x \in X} \twoNormSqr{c_i^* - x} -\frac{1}{|Y|}  \sum_{y \in Y} \twoNormSqr{y - c_j^*} \leq  \frac{1}{|X||Y|} \sum_{x \in X} \sum_{y \in Y} \twoNormSqr{x - y} 
 \end{align*}}\vspace{-2mm}
 Since $\twoNormSqr{c_i^* - c_j^*} \geq \delta \cdot R$ and by the definition of $R$, 
{\tiny \begin{align*}
 (\frac{1}{3}\delta-2) \cdot R \leq \frac{1}{3}\twoNormSqr{c_i^* - c_j^*} - \frac{1}{|X|}  \sum_{x \in X} \twoNormSqr{c_i^* - x} -\frac{1}{|Y|}  \sum_{y \in Y} \twoNormSqr{y - c_j^*} \leq \frac{1}{|X||Y|} \sum_{x \in X} \sum_{y \in Y} \twoNormSqr{x - y}  
\end{align*}\vspace{-2mm}
 }
 
However, for any two subclusters $X, X' \subset C^*_i$ we know that:
{\tiny \begin{align*}\vspace{-2mm}
\frac{1}{|X||X'|} \sum_{x \in X} \sum_{x \in X'} \twoNormSqr{x - x'} \leq   2 \left ( \frac{1}{|X|} \sum_{x \in X} \twoNormSqr{c^*_i - x} +  \leq \frac{1}{|X'|} \sum_{x \in X'} \twoNormSqr{c^*_i  - x'} \right ) \leq 4 R.
\end{align*}}

For $\delta \geq 30$, we know that there exists
a $4R \leq \tau_r \leq 8R$ for which $X$ and $X'$ would form a sub-cluster component without $Y$. Since we use a geometric sequence of $\tau_1,\tau_2,\dots$, we know that $\tau_r$ will exist since it is between $4R$ and the doubling of $4R$. $X$ and $X'$ are any sub-clusters of any ground truth cluster $C_i^*$. 
The result above indicates that at any round before the one using $\tau_r$ that takes as input pure sub-clusters will produce pure sub-clusters as no sub-clusters with points belonging to different ground truth clusters will be merged.
Moreover, the existence of $\tau_r$ indicates that the partition given by sub-cluster component from a round using $\tau_r$ , will contain every ground truth cluster in $\clustering^*$. In particular, the last round that uses $\tau_r$ will be the $r^\star$ to do this,  i.e., $\clustering^{(r^\star)} = \clustering^\star$. Observe that the separation condition requires that within cluster distances for any two subsets will be less than $\tau_r$ and so sub-clusters will continue to be merged together until each ground truth cluster is formed by the last round using $\tau_r$.

\subsection{Proof of Proposition \ref{prop:mbs}}
\label{proof:mbs}

Previous work shows that hierarchical
agglomerative clustering (HAC) can
recover model-based separated data \cite{monath2019scalable}. Proposition \ref{prop:hac} shows that there exists
a sequence of thresholds for which SCC 
produces the same tree structure as HAC 
and thus this same sequence of thresholds 
could be used to recover model-based separate data. It would also be possible to compress this sequence of thresholds to contain only those thresholds necessary to prevent the overmerging of clusters with points from different ground truth classes (which must exist by the previous result).

\subsection{Proof of Theorem \ref{thm:dpmeans}}

We showed that under $\delta$-separation
\scc will recover the target partition (Theorem~\ref{thm:opt}). 
We first relate this target partition to the facility location using the DP-Facility Problem and then use the relationship between Facility Location and DP-Means \citep{pan2013optimistic}. Facility Location problem is defined as:

\begin{definition}\textbf{\emph{[Facility Location]}}
Given a set of clients $H$ as well as facilities $F$, and a set 
$\fb = \{f_1, \dots,  f_K\}$ of  facility opening costs, that is let $f_j$ be the cost of opening facility $j$ and let $\fc(i,j)$ be the cost of 
connecting client $i$ to the open facility $j$. Let $I \subseteq F$ be the set of 
opened facilities and let $\phi: H \rightarrow I$ be the mapping from clients to facilities. 
The total cost of opening a set of facilities is: 
\begin{align}
    cost(H, F, I, \phi, \fb) = \sum_{i \in H} \fc(i, \phi(i)) + \sum_{i \in I} f_i
\end{align}
The facility location problem is to solve: 
$\argmin_{I, \phi} cost(H, F, I, \phi, \fb)$ given $\fc$, $F$ and $\fb$.
\end{definition}

\noindent As shown by \citet{pan2013optimistic}, Facility Location is closely related to DP-Means. In particular, the solution of Facility Location gives a solution to DP-Means:

\begin{definition}\textbf{\emph{[DP-Facility \citep{pan2013optimistic}]}}  
We define the DP-Facility problem to be the facility location problem where 
$f_j = \lambda$ for all facilities $j \in F$, $H=F=\dataset$ and $\fc$ be 
\emph{squared euclidean distance}. Given a solution $I, \phi = \argmin_{I, 
\phi} cost(\dataset, \dataset, I, \phi, \mathbf{\lambda})$, we define that $c_. \defeq I$ and 
$C_k = \{ i | \phi(i) = x_k \}$, $K=|I|$ and say that $\clustering = (C_1, \ldots, C_K)$ is a solution to DP-Means given by 
the solution of DP-Facility.
\end{definition}

First, we consider $\delta$-separated data in DP-Facility problem and show that the target separated partition gives an optimal solution to DP-Facility:

\begin{proposition}\label{thm:fclty}
Suppose the dataset $\dataset$ satisfies the $\delta$-separability assumption with 
respect to clustering  $C^\star_1,\dots,C^\star_k$,   then this clustering is an optimal solution  
to the DP-Facility problem with $\lambda = (\delta-2) \cdot R$. 
where $R:=\max_{l \in [k]} \max_{x  \in C^\star_l} \norm{x - c_l^*}$. 
\end{proposition}
\begin{proof}[Proof]
To show that this clustering is 
an optimal solution to the DP-Facility problem, we will 
use linear programming duality.  In particular, we will exhibit a feasible dual, $\alpha$,
to the linear programming 
relaxation of the DP-Facility Problem, whose cost is the same as the clustering 
$\{ C^*_1, \dots, C^*_k \}$.  From linear programming duality, we know that the following
set of relations are true :
$\textsc{cost}(\alpha) \leq \textsc{Opt(Dual)} = \textsc{OPT(Primal)} \leq \textsc{cost}(C^*)$. 
Combined with the fact that $\textsc{cost}(\alpha) = \textsc{cost}(C^*)$,  this will show that
clustering is an optimal solution to the DP-Facility problem. 

Consider the  linear programming relaxation to  DP-Facility problem. This LP is 
an adaption of the classical LP used for the facility location problem
considered in~\cite{vazirani2013approximation}[Ch.~17]. 
\begin{align*}
\tiny
    \min &\sum_{i \in F}  \sum_{j \in C} \fc(i, j) \cdot z_{i,j} + \lambda \sum_{i \in F} y_i  \\
     & \quad \sum_{i \in F} z_{ij} \geq 1 \qquad  \text{ for all } j \in C \\
     &\quad  y_i  - z_{ij} \geq 0    \qquad \text{ for all  } j \in C \\
     &\quad z,y \geq 0 
\end{align*}
The above program contains two variables $z_{ij}$ indicating if client $j$ is 
connected to facility $i$ and variables $y_i$ indicates if facility $i$ is open. In particular, every feasible solution to the DP-Facility problem is a candidate solution to the above LP. 
The dual to the above program is given below:
\begin{align*}
\tiny
    \max \sum_{j \in C} \alpha_j \\
    \sum_j \beta_{ij} &\leq \lambda \qquad  \text{ for all $i \in F$} \\ 
    (\alpha_j - \fc(i,j) ) &\leq \beta_{ij}  \qquad \text{for all }i \in F, j \in C \\
    \alpha, \beta \geq 0
\end{align*}

For each cluster $C_i$, and each  point in the cluster $x \in C_i$,  
$\alpha_{x} = ((\delta-2)R + \fc(c^*_i,x)/r$  where $r$ is the size of cluster 
$r:=|C_j|$.   By $\delta$-separability assumption, we can deduce that  
$\beta_{\tilde{i}x} = 0$ for all other clusters $C^*_{\tilde{i}} \neq C^*_i$. However,
for all $x \in C_i$, we will have $\sum_{x \in C_i} \beta_{ix} = r\cdot \frac{\lambda}{r} = \lambda $. This shows that $\alpha$ is a valid dual to the LP. 
\end{proof}

The next proposition formally relates the DP-Facility problem to an approximate solution to  
the DP-Means objective.
\begin{proposition} \citep{pan2013optimistic}
\label{prop:fl2dp}
Let $\mub^\star,Z^\star, K^\star$ be an optimal solution to the DP-Means problem and let 
$\mub^\dagger,Z^\dagger, K^\dagger$ be the DP-Means solution given by an optimal solution, $I^\dagger,\phi^\dagger$ to the DP-facility location problem. Then, 
$DP(X, \lambda, Z^\dagger, \mub^\dagger, K^\dagger) \leq 2 \cdot DP(X, \lambda, Z^*, \mub^*, K^*)$
\end{proposition}

 Finally, we can analyze the quality of the solutions found by \scc on $\delta$-separated data, showing that it is a constant factor approximation. Using theorem \ref{thm:opt} and Proposition \ref{thm:fclty}, when the data satisfies the $\delta-$separation assumption, then \scc contains the optimal solution to DP-facility problem. Finally using Proposition \ref{prop:fl2dp}, we see that this solution is within $2$ factor of the DP-means solution.

\begin{figure}
  \begin{center}
    \includegraphics[width=0.35\textwidth]{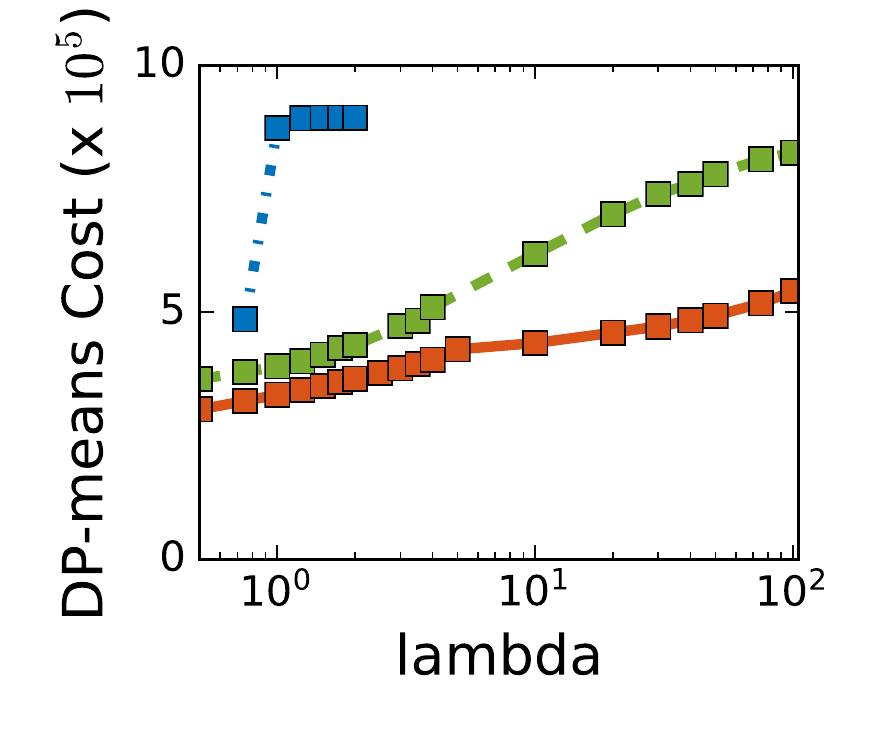}
    \includegraphics[width=0.35\textwidth]{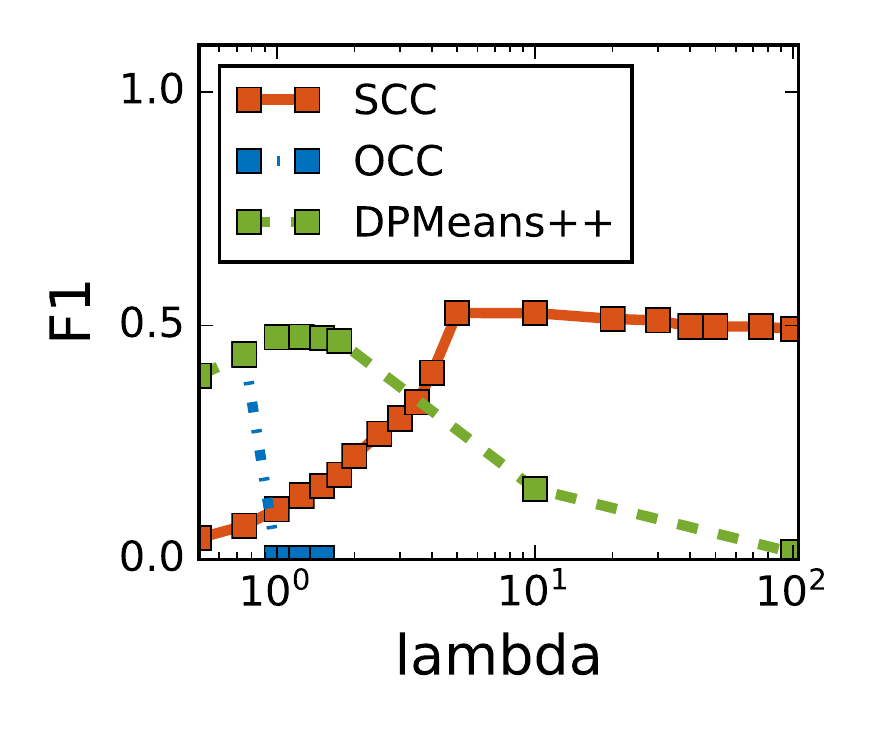} 
  \end{center}
  \caption{DP-Means and F1 accuracy for ILSVRC (Lg.) dataset}
\end{figure}
\clearpage
\section{Experiments}
\label{sec:app_experiments}

\textbf{ILSVRC (Lg.) DP-Means}. We replace the SerialDPMeans
baseline with its parallel/distributed variant \textbf{OCC} \citep{pan2013optimistic}.
OCC can only be run with $\lambda<4.0$
(max normalized $\ell_2^2$ distance). We find that it takes longer than 10 hours to run more than 2 iterations of OCC,
for lambda values less than 0.75. After these iterations, we observe that the lambda value of 0.75 gives a reasonable F1 score for OCC. 
We observe that \scc achieves higher F1 scores when the value of $\lambda$ is larger, as having
too small a $\lambda$ value creates too many clusters. We observe that \scc produces
lower costs than DPMeans$++$ and achieves a higher F1 value for particular values of $\lambda$.

\textbf{LowrankALBCD \citep{yen2016scalable}} uses values of lambda at the scale of $\lambda= 0.01 \cdot N$ and normalized $\ell_2^2$.
Any value of $\lambda$ greater than the maximum pairwise distance (4)
will result in SerialDPMeans and OCC necessarily giving solutions of every data point in the same cluster. We evaluated LowrankALBCD on CovType, ALOI, Speaker, and ILSVRC (Small). We use the code provided by the authors of LowrankALBCD to solutions with small values of $\lambda$ in the range 0 to 4, for large numbers of iterations we found the code either required more than 100GB of RAM or took longer than 10 hours. Instead, we compare our method and LowrankALBCD for larger values of lambda, specifically the authors suggestion of $0.01N$. We observe that 
on \scc and LowrankALBCD perform similarly on several of the datasets and LowrankALBCD performs better on CovType. However, we notice that for all datasets except
CovType, these values of $\lambda$ produce unreasonably few clusters.

 \textbf{Running Times}. We report running times in Table~\ref{tab:runningTimes}. The time
required by \alg and Affinity is dominated by the construction
of the sparse nearest neighbor graph. We use the publicly available implementations of HDBSCAN and Grinch.

\begin{table}
    \centering
    \begin{tabular}{l@{}ccc}
    \toprule
         &  \bf ALOI & \bf ILSVRC (Sm.)  & \bf ILSVRC (Lg.)  \\
         \midrule 
         \textsc{HDBSCAN} & 1635 & 7902.90 & DNF \\
         \textsc{Grinch} & 385.113 & 836.748 & DNF \\
         \textsc{Affinity} & 29.01 + 0.834 & 261.831 + 0.298 & 849.846 + 9.886 \\
        \alg & 29.01 + 2.79 & 261.831 + 4.29  & 849.846 + 65.621 \\
        \bottomrule
    \end{tabular}
    \caption{\textbf{Running Time (seconds) of Top Performing Methods}. Each run on machine using 24 2.40GHz CPUs. Grinch and HDBSCAN did not finish on largest dataset in 10 hours. Affinity and SCC report Sparse Graph Construction Time + Algorithm Execution Time for given graph. DNF = Did not finish in 10hours.}
    \label{tab:runningTimes}
\end{table}

\textbf{Robust Hierarchical Clustering (RHC) \cite{balcan2014robust}} RHC has stronger theoretical guarantees than \scc. RHC achieves these stronger theoretical guarantees by using a complex linkage function, requiring more time per linkage function execution. In Table~\ref{tab:rhc}, we compare the best dendrogram purity achieved by SCC and RHC on the Iris and Wine datasets using a grid search over each method's hyperparameters ($\alpha+\nu$ for RHC, number of nearest neighbors and rounds for \scc). We use the publicly available MATLAB implementation of RHC. We find on these small datasets that \scc achieves competitive dendrogram purities despite using a simpler linkage function.

\begin{table}
\centering
\begin{tabular}{lcc}
\toprule
     & RHC & SCC \\
\midrule
Iris & \bf 0.955  &           0.926 \\
Wine &  0.944 &           \bf 0.975 \\
\bottomrule
\end{tabular}
\caption{\textbf{Comparison to RHC} We report the best dendrogram purity achieved across various hyperparameter settings of each method.}
\label{tab:rhc}
\end{table}

\end{document}